\documentclass{article}

\usepackage{geometry}
\usepackage{amsmath, amssymb}
\usepackage{authblk}
\usepackage{graphicx}
\usepackage{hyperref}
\usepackage{capt-of}
\usepackage{algorithm}
\usepackage{algpseudocode}
\usepackage{booktabs}
\usepackage{multirow}
\usepackage{placeins}

\usepackage[style=apa, sorting=nyt]{biblatex}
\newtheorem{remark}{Remark}

\title{Does a Large Language Model Really Speak in Human-Like Language?}

\author[1]{Mose Park}
\author[1]{Yunjin Choi}
\author[1]{Jong-June Jeon}

\affil[1]{Department of Statistical Data Science, University of Seoul, Seoul, South Korea}

\date{}

\begin{document}

\maketitle

\begin{abstract}
Large Language Models (LLMs) have recently emerged, attracting considerable attention due to their ability to generate highly natural, human-like text. This study compares the latent community structures of LLM-generated text and human-written text within a hypothesis testing procedure. Specifically, we analyze three text sets: original human-written texts ($\mathcal{O}$), their LLM-paraphrased versions ($\mathcal{G}$), and a twice-paraphrased set ($\mathcal{S}$) derived from $\mathcal{G}$. Our analysis addresses two key questions: 
(1) Is the difference in latent community structures between $\mathcal{O}$ and $\mathcal{G}$ the same as that between $\mathcal{G}$ and $\mathcal{S}$? (2) Does $\mathcal{G}$ become more similar to $\mathcal{O}$ as the LLM parameter controlling text variability is adjusted? The first question is based on the assumption that if LLM-generated text truly resembles human language, then the gap between the pair ($\mathcal{O}$, $\mathcal{G}$) should be similar to that between the pair ($\mathcal{G}$, $\mathcal{S}$), as both pairs consist of an original text and its paraphrase. The second question examines whether the degree of similarity between LLM-generated and human text varies with changes in the breadth of text generation. To address these questions, we propose a statistical hypothesis testing framework that leverages the fact that each text has corresponding parts across $\mathcal{O}$, $\mathcal{G}$, and $\mathcal{S}$. This relationship enables the mapping of one dataset's relative position to another, allowing two datasets to be mapped to a third dataset. As a result, both mapped datasets can be quantified with respect to the space characterized by the third dataset, facilitating a direct comparison between them. For $\mathcal{O}$, the original human text, we collected customer reviews from an accommodation booking site; for $\mathcal{G}$ and $\mathcal{S}$, we used GPT-3.5 to paraphrase $\mathcal{O}$. Our results indicate that GPT-generated text remains distinct from human-authored text.
\end{abstract}

\section{Introduction}\label{sec1}

Recently, large language models (LLMs) have attracted significant attention for their remarkable performance across various natural language processing (NLP) tasks \parencite{zhao2024surveylargelanguagemodels}. These models have demonstrated their capabilities beyond traditional text classification, achieving high accuracy in machine translation, question answering, and summarization tasks \parencite{edge2024local, bang2023multitask}. Furthermore, current research efforts focus on assessing the reasoning abilities of LLMs or deploying them as agents capable of replicating human-like behavior \parencite{ramezani2023knowledge, wang2024probing}. The reasoning capabilities of LLMs extend beyond simple language processing to include an understanding of context and logical structures, enabling appropriate responses in complex, interactive scenarios \parencite{guo2023evaluatinglargelanguagemodels}. Additionally, research into LLMs as autonomous agents or "virtual clones" that can undertake specific tasks independently is expanding, showing a shift toward developing LLMs as intelligent agents capable of understanding and mimicking human intentions and preferences \parencite{kim2024debate}. As a result, LLMs are expanding the field of NLP, finding applications across various domains.

Despite these advancements, the field still lacks robust statistical approaches for quantitatively assessing the features of LLMs, largely due to their recent introduction.
Some pioneering studies have focused on distinct aspects of LLM evaluation. 
\textcite{chiba2024tackling} explored statistical methods to estimate the originality of GPT-generated content in comparison to existing content subject to copyright. 
\textcite{jiang2024peektokenbiaslarge} proposed a statistical hypothesis testing approach to examine how socially sensitive input tokens, such as those related to gender or ethnicity, influence LLM-generated answer which are LLM-generated text.
\textcite{cherian2024large} proposed a method to enhance the validity of LLM-generated texts using conformal prediction methods.

Also, statistical approaches have been introduced to detect LLM-generated texts by investigating their watermarks \parencite{li2024statistical, xie2024debiasing, li2024robust}.
While these studies address important topics, they do not consider how closely the LLM-generated texts resemble actual human-authored text, as this aspect lies outside their scope.  
Moreover, statistical tools for testing the differences between  LLM-generated text and real discourse remain underdeveloped.
Specifically, real discourse often exhibits greater lexical richness and diversity than LLM-generated text, suggesting a possible gap in linguistic complexity \parencite{martinez2024beware}. For instance, previous research has revealed distinct linguistic patterns between human-authored news texts and those generated by LLMs, further indicating the need for more comprehensive statistical analyses to explore these differences in depth \parencite{munoz2024contrasting}.

\begin{figure*}[t]
    \centering
        \includegraphics[width=0.75\textwidth]{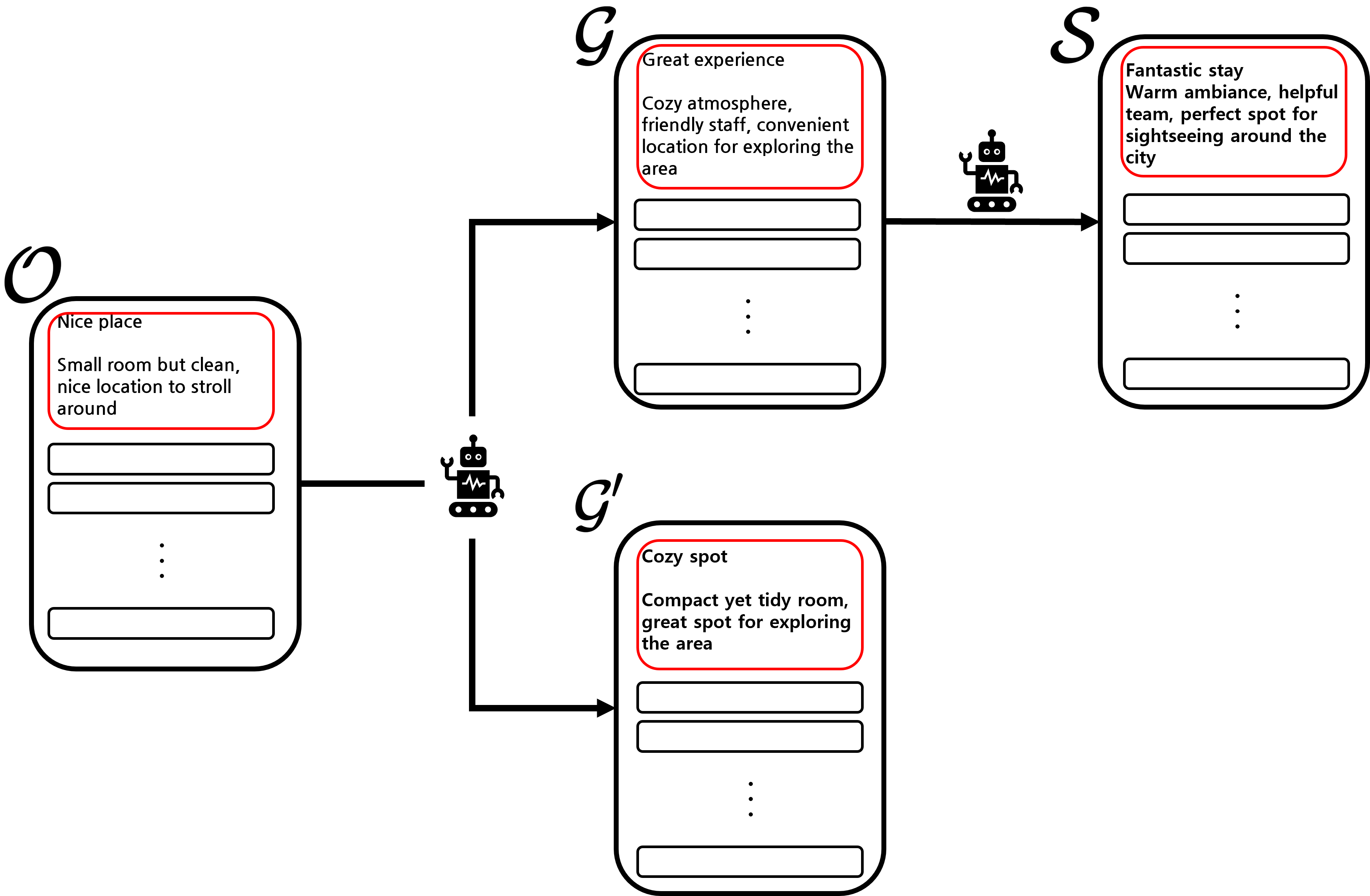}
        \caption{Text generation framework }\label{fig1:text_generation}
\end{figure*}

This study investigates the measurable gap between human-authored and LLM-generated expressions using a focused statistical approach. 
The key contribution of our work lies in proposing a comprehensive framework that spans the entire process—from data collection and problem formulation to the development of a hypothesis testing method. Each step is specifically designed to address the question of whether LLM-generated text is similar to human-authored text, taking into account the characteristics of our collected data.
Building on this foundation, we make two specific advances. First, we use a statistical hypothesis testing framework to investigate the differences in the latent community structures between human-authored and LLM-generated text. Second, to facilitate this analysis, we propose a statistical hypothesis testing procedure for relative differences, with the reference standard specified by the user.
For the statistical analysis, we collected over 30,000 human-authored texts through web crawling. We then generated LLM-produced texts paired with the collected human-authored texts by creating paraphrases of the original inputs. The LLM-generated data were produced multiple times under various settings. 
We also used three additional datasets of human-authored texts (CNN, SQuAD2, and Quora) provided in \textcite{chatgpt_paraphrases_dataset}, for which LLM-generated paraphrases were similarly created.
Our proposed hypothesis testing procedure utilizes the paired data structure between the original texts and their paraphrases,  based on their embeddings.
Specifically, this method leverages the relative positions of text embeddings in one dataset compared to those in another, as each text has a corresponding pair in the other dataset. 
As a result, when multiple datasets share this relationship, they can be mapped to one of the datasets and quantified with respect to the space characterized by the dataset that serves as the reference, facilitating direct comparison across distinct datasets.
This approach therefore circumvents the challenge of directly comparing the community structures of different text datasets, as it only needs to account for variability relative to a single dataset.

\subsection{Related Work}\label{subsec2}

Our study aims to investigate the differences between human-authored and LLM-generated texts using statistical hypothesis testing. This approach is closely related to hypothesis testing in NLP tasks, as it involves text data. In the field of NLP, hypothesis testing has been widely applied, primarily to evaluate the performance of NLP outputs. 
For example, \textcite{lewis2022statistical} used statistical hypothesis testing to assess the goodness of fit of a topic model based on word frequency.

\textcite{dror2017replicability} focused on reproducibility analyses in NLP performance evaluation, examining performance across multiple datasets and refining statistical validation by addressing multiple comparison issues.

Previous studies have also focused on appropriately handling natural language data, as it differs from traditional data formats used for statistical hypothesis testing and presents additional challenges.

Several studies have revisited the assumptions and limitations of existing evaluation methodologies \parencite{riezler2005some, smucker2007comparison} and provided guidelines for metric selection and significance testing \parencite{dror2018hitchhiker, koplenig2019against}.

Also, to address the non-standard distribution issue in linguistic data, some studies have employed nonparametric hypothesis testing methods to evaluate NLP outputs.

For instance, \textcite{koehn2004statistical} applied bootstrap testing in machine translation, while \textcite{urbano2019statistical} utilized Wilcoxon signed-rank tests, permutation tests, and bootstrap tests in information retrieval.

Additionally, \textcite{berg2012empirical} and \textcite{dror2020statistical} extended bootstrap testing to diverse NLP tasks, including machine translation, summarization, and parsing.

In our study, we similarly employ a nonparametric hypothesis testing method to address distributional challenges in linguistic data.

In that our study performs statistical hypothesis testing to detect differences between two text datasets, it shares similarities with the studies of \textcite{deng2021two} and \textcite{liu2021statistically}. Specifically, \textcite{deng2021two} examined relatively long texts, such as novels, by testing differences between the initial and later sections of the same text in terms of word frequency and vocabulary diversity using the Wilcoxon signed-rank test. This study explores how linguistic characteristics within a single text can vary over time or in different positional contexts by comparing distinct intervals within the same document. 

In contrast, our study uses text embedding space to capture potential differences in content, rather than focusing on word usage such as frequency or diversity.

\textcite{liu2021statistically} is more closely aligned with our study in that it also utilizes embedding space. 

Specifically, this study uses contextual token embeddings derived from BERT to detect semantic changes in texts and verifies the statistical significance of these differences using a permutation test. 

While both \textcite{liu2021statistically} and our study share a focus on using embedding space to capture subtle differences in contents, our approach differs in that it examines differences in the community structure of the data, whereas \textcite{liu2021statistically} focuses on detecting semantic shifts.

Additionally, our study differs from both of these studies in that we compare texts from two distinct sources (human versus LLM), which are expected to share the same meaning in that the LLM generates paraphrases of the human-authored text. This paired structure plays a crucial role in our hypothesis testing procedure.

Our study aligns with the approaches proposed in \textcite{li2024statistical, xie2024debiasing, li2024robust}, which aim to distinguish between LLM-generated and human-authored texts within a statistical hypothesis testing framework. As LLMs are capable of generating highly realistic texts, notable socioethical concerns have emerged, including the creation of hard-to-detect fake news and malicious content \parencite{ferrara2024genai, wu2023survey}. To address these concerns, it has been proposed that LLM-generated texts incorporate nearly unnoticeable signals, referred to as watermarks.
Building on these watermarks, aforementioned studies investigated methods for detecting LLM-generated texts.

Specifically, \textcite{li2024robust} introduced a robust hypothesis testing approach using a truncated goodness-of-fit test, which remains effective even when the text has been edited by humans. \textcite{li2024statistical} proposed a statistical framework for designing rigorous watermark detection rules by precisely evaluating Type I and Type II errors. \textcite{xie2024debiasing} developed a debiasing scheme for watermarking, analyzing it from the perspective of sparse signal detection. All these methods focus on determining whether a given text was generated by an LLM when only the text itself is provided. On the other hand, our study examines the differences in the latent community structures between LLM-generated and human-authored texts, using the actual texts from both sources, where both convey the same underlying meaning.

The remainder of the paper is organized as follows: Section \ref{sec:overview} provides an overview of the study; Section \ref{sec:data_description} describes the datasets used, including human-authored text and LLM-generated text; Section \ref{sec:hyp} outlines the hypothesis testing procedure; and Section \ref{sec:res} presents the results of the data analysis. The paper concludes with a discussion.

\section{Overview \label{sec:overview}}

The primary focus of our study is to investigate whether LLM-generated data is indeed similar to human-authored text. To address this issue, we examine four distinct types of datasets and propose a hypothesis testing method based on the structure of these datasets: human-authored text ($\mathcal{O}$), LLM-generated paraphrases of human text ($\mathcal{G}$), and paraphrases of LLM-generated datasets, which are themselves generated using LLM ($\mathcal{S}$). Specifically, when the \( i \)-th text in \( \mathcal{O} \) is input into the LLM for paraphrasing, the resulting paraphrase is the \( i \)-th text in \( \mathcal{G} \). Likewise, when the \( i \)-th text in \( \mathcal{G} \) is input into the LLM for paraphrasing, the resulting output is the \( i \)-th text in \( \mathcal{S} \). Thus, the \( i \)-th observations across the datasets are paired.
In addition to \( \mathcal{G} \), we consider another paraphrased text dataset generated using the LLM from \( \mathcal{O} \), denoted as \( \mathcal{G}' \).
Figure \ref{fig1:text_generation} illustrates the text generation framework.
Based on these datasets, we aim to answer the following two questions:
\begin{enumerate}
    \item[Q1.] Are the latent community structures of $\mathcal{O}$ and $\mathcal{G}$ (and, equivalently, $\mathcal{G'}$) identical?
    \item[Q2.] Does the latent community structure of $\mathcal{G}$ become similar to that of $\mathcal{O}$ as the LLM parameters controlling text variability are adjusted?
\end{enumerate}
We propose a class of hypotheses corresponding to each question, aimed at addressing them, and introduce the rationale behind the proposed procedure. The hypotheses introduced in this section are informal, as some terms have not been formally defined yet. However, we present them this way to emphasize the main ideas. The formal hypothesis is presented in Section \ref{sec:hyp}.

\subsection{Hypothesis Design for Question One \label{subsec:hypq1}}

The first question is based on the assumption that if LLM-generated text truly resembles human language, the latent community structures of \( \mathcal{O} \) and \( \mathcal{G} \) will be similar. This is particularly plausible since \( \mathcal{G} \) is generated based on \( \mathcal{O} \), implying that each text in the datasets is paired. In this case, the difference in the community structures between the pair \( (\mathcal{O}, \mathcal{G}) \) is expected to closely align with that of the pair \( (\mathcal{G}, \mathcal{S}) \), where the latter pair consists entirely of LLM-generated texts and is therefore likely to exhibit similar community structure.
Conversely, if the LLM-generated texts differ from human-authored text, \( \mathcal{O} \) and \( \mathcal{G} \) would have distinct latent community structures, while those of \( \mathcal{G} \) and \( \mathcal{S} \) would be closely aligned.
In this scenario, the gaps in community structures between the \( (\mathcal{O}, \ \mathcal{G} )\) pair and that between \( (\mathcal{G}, \  \mathcal{S} ) \) pair would not be symmetric.
Based on these assumptions, to address the first question, we  investigate the difference in community structures between the \( (\mathcal{O}, \  \mathcal{G} ) \) pair, using that of the \( (\mathcal{G}, \  \mathcal{S} ) \) pair as a reference.
This is examined through the following null hypothesis, which is likely to be rejected if the latent community structure of \( \mathcal{O} \)  and \( \mathcal{G} \) are distinct:
\begin{eqnarray*}
     H_{0} (\mathcal{G}, \{\mathcal{O}, \mathcal{S}\}) &:& \text{The gap in community structure between } (\mathcal{G}, \mathcal{O}) \\
     & & \text{and that between } (\mathcal{G}, \mathcal{S}) \text{ are the same.}
\end{eqnarray*}
In the same vein, we also investigate the gap between the pairs of $(\mathcal{O}, \ \mathcal{G})$ and $(\mathcal{O}, \ \mathcal{G}')$ 
by examining the following null hypotheses:

\begin{eqnarray*}
     H_{0} (\mathcal{G}, \{\mathcal{O}, \mathcal{S}\}) &:& \text{The gap in community structure between } (\mathcal{G}, \mathcal{O}) \\
     & & \text{and that between } (\mathcal{G}, \mathcal{S}) \text{ are the same.}
\end{eqnarray*}

Similar to the aforementioned discussion, in testing the relationships among \( \mathcal{O} \), \( \mathcal{G} \), and \( \mathcal{G}' \), \({H}_{0}(\mathcal{G}, \{\mathcal{O}, \mathcal{G}'\} )\) is expected to be rejected when \( \mathcal{O} \) and \( \mathcal{G} \) have different latent community structure. 

\begin{remark}
     The null hypothesis ``$H_{0} (\mathcal{O}, \{\mathcal{G}, \mathcal{S}\})$'' which tests ``The gap in community structure between   $(\mathcal{O}, \mathcal{G})$ and that between  $(\mathcal{O}, \mathcal{S})$  are the same.'' is likely to be rejected even when the community structure of $\mathcal{G}$ is close to that of $\mathcal{O}$. This is because the gap between \( (\mathcal{O}, \mathcal{G}) \) is expected to consistently be smaller than that between \( (\mathcal{O}, \mathcal{S}) \) as
     \( \mathcal{O} \) and \( \mathcal{G} \) are one-step paraphrases, while \( \mathcal{O} \) and \( \mathcal{S} \) are two-step paraphrases, with \( \mathcal{G} \)  acting as a bridge between \( \mathcal{O} \) and \( \mathcal{S} \).
Thus, the results of testing $H_{0} (\mathcal{O}, \{\mathcal{G}, \mathcal{S}\})$ are presented in Section \ref{sec:res} to validate the power of the proposed hypothesis testing method.
\end{remark}

\begin{remark}
Since \( \mathcal{G} \) and \( \mathcal{G}' \) are generated in parallel, switching their roles should not affect the test results. Therefore, the null hypothesis 
``\( H_{0} (\mathcal{G}', \{\mathcal{O}, \mathcal{G}\}) : 
\text{The gap in community structure between } (\mathcal{G}', \mathcal{O}) \\
\text{and that between } (\mathcal{G}', \mathcal{G}) \text{ are the same.} \)`` is omitted as it is redundant.
Also, as the community structures of \( \mathcal{G} \) and \( \mathcal{G}' \) are equivalent, we also expect ``\( H_{0} (\mathcal{O}, \{\mathcal{G}, \mathcal{G}'\}) : 
\text{The gap in community structure between } (\mathcal{O}, \mathcal{G}) \\
\text{and that between } (\mathcal{O}, \mathcal{G}') \text{ are the same.} \)'' to be accepted in all cases. Thus, the results of testing \( H_{0} (\mathcal{O}, \{\mathcal{G}, \mathcal{G}'\}) \) are presented in Section \ref{sec:res} to further validate the effectiveness of the proposed hypothesis testing method.

\end{remark}

\subsection{Hypothesis Design for Question Two \label{subsec:hyptwo}}

The second question is based on the assumption that the degree to which LLM-generated text resembles human-authored text may depend on a specific parameter known as the temperature parameter\footnote{https://platform.openai.com/docs/api-reference/audio/createTranscription}, which controls the variability of the generated texts. Specifically, as the results of text generation vary—-particularly in terms of breadth—based on the temperature parameter, it is possible that at certain settings, the latent community structure of generated texts may align more closely with human writing.
Below is an example of an original text and its paraphrased result using an LLM at various temperature settings, illustrating how the output varies with the temperature parameter.
\begin{itemize}
    \item An original user review. \\
    ``{\it Basic, clean and comfortable hotel. Not bad for a short stay. The accessibility to everything is great. }''
    \item The paraphrased result at a temperature parameter of 0.1. \\
``{\it A budget-friendly hotel that offers clean and cozy accommodations. Perfect for a quick stay. The convenient location makes it easy to access all amenities. }''
    \item The paraphrased result at a temperature parameter of 0.7. \\
    ``{\it Cozy, well-maintained hotel with all the essentials. Perfect for a quick getaway. The convenience of its location is a huge plus.}''
    \item The paraphrased result at a temperature parameter of 1.5. \\
    ``{\it This boutique hotel exceeded my expectations. The room was cozy and the staff was friendly. The location was perfect for exploring the city. A great choice for a quick getaway.}''
\end{itemize}

At the low temperature of 0.1, the generated reviews closely adhere to the original tone, incorporating key words such as `clean,' `comfortable,' and `accessible.'
At the medium temperature of 0.7, which is the default setting, the generated text introduces a bit more descriptive richness and natural language variation. In this example, words like `cozy' and `well-maintained' are used more liberally.
At the high temperature of 1.5, the reviews become notably more expressive, with shifts in tone. Phrases such as `exceeded my expectations' and `a great choice for a quick getaway' make the text feel more enthusiastic compared to the original. 
This example suggests that a temperature setting that is too high may alter the input text excessively, resulting in a significant disparity between $\mathcal{O}$ and  $\mathcal{G}$. Conversely, a temperature setting that is too low may struggle to paraphrase the input with minimal changes, leading to outputs that lack human-like qualities. 
This indicates that the degree to which the generated text resembles human-authored text may depend on the temperature parameter.

To address this question, we examine comparing generated texts with different temperatures by examining the following hypothesis:

\begin{eqnarray*}
     H_{0} (\mathcal{G}_{\rho_1}, \ \{\mathcal{O}, \ \mathcal{G}_{\rho_2}\}) &:& 
     \text{The gap in community structure between } (\mathcal{G}_{\rho_1}, \mathcal{O}) \\
     & & \text{and that between } (\mathcal{G}_{\rho_1}, \mathcal{G}_{\rho_2}) \text{ are the same.}
\end{eqnarray*}

for distinct $\rho_1$ and $\rho_2$ values where $\mathcal{G}_{\rho}$ represents the paraphrased text generated by the LLM at the temperature parameter \(\rho\).
While the degree of difference between the pair \( (\mathcal{O}, \mathcal{G}_{\rho_2}) \) is not directly available, that between the pair \( (\mathcal{G}_{\rho_1}, \mathcal{G}_{\rho_2}) \) can be quantified by the difference in the temperature parameter \( \left|\rho_1 - \rho_2\right| \). Thus, the proposed null hypothesis $ H_{0} (\mathcal{G}_{\rho_1},  \ \{\mathcal{O}, \ \mathcal{G}_{\rho_2}\}) )$ uses the pair \( (\mathcal{G}_{\rho_1}, \mathcal{G}_{\rho_2}) \) as a reference to measure the difference between the pair \( (\mathcal{O}, \mathcal{G}_{\rho}) \) by comparing the two pairs.

\subsection{The Rationale Behind the Proposed Hypothesis Testing Procedure \label{subsec:teststat}}
All the hypotheses we examine involve assessing whether the disparity between the gap of one pair of text sets differs from the gap of another pair, where both pairs share one dataset in common as an anchor. This is analogous to comparing the lengths of two sides of a triangle.
For instance, the hypothesis \( H_{0} (\mathcal{G}, \{\mathcal{O}, \mathcal{G'}\})\), with \(\mathcal{G}\) as an anchor, is analogous to comparing the lengths between \( (\mathcal{G}, \mathcal{O})\) and \( (\mathcal{G}, \mathcal{G}')\) pairs, where \( \mathcal{O}\), \(\mathcal{G}\), and \(\mathcal{G'}\) are the vertices of a triangle.

While comparing the distributions between \( \mathcal{O} \) and \( \mathcal{G} \) may seem more direct than examining their latent community structures, it is challenging due to the nature of text data in that it lacks a common standardized quantification, is high-dimensional, and may exist in different spaces, among other complications.  
Thus, our approach is designed to circumvent these difficulties by focusing on the community structure of the datasets, based on the idea that if the datasets are distributionally the same, their community structures should also align.
Our formal hypothesis, introduced in Section \ref{sec:hyp}, is based on the fact that the datasets are paired, ensuring a one-to-one mapping between them.
In this setting, the question of whether two communities are identical simplifies to whether two partitions of a given index set are identical, as the two datasets share the same index set.
In the testing procedure, our approach again takes advantage of the paired nature of the datasets, ensuring that each data point in a non-anchor dataset has a corresponding point in the anchor dataset. Consequently, this one-to-one mapping allows any statistic calculated on the non-anchor data to be mapped to the anchor data.
This facilitates the comparison of the two statistics, each respectively derived from the two non-anchor datasets, as both mapped statistics are quantified in the same space defined by the anchor dataset and share the same unit of measurement. Specifically, our procedure derives the statistic from clustering applied to each of the two datasets, based on the premise that their clustering results will align when the latent community structures of the datasets are the same, resulting in two mapped statistics with  similar values.

The outline of the testing procedure is as follows:

\begin{description}
    \item[Step 1.] Perform clustering on the two non-anchor datasets on their embeddings respectively.
    \item[Step 2.] 
    Map the distances from the observations to their corresponding cluster centers in the two datasets onto the anchor dataset, respectively.
    \item[Step 3.] Perform hypothesis testing to examine whether the location parameters of the two mapped distances are the same.
\end{description}

\section{Data Description \label{sec:data_description}}
This section provides a description of the actual data utilized in our analysis.
\subsection{Collecting Original Human Text \texorpdfstring{$\mathcal{O}$}{O}}
\begin{figure*}[t]
    \centering
        \includegraphics[width=0.75\textwidth]{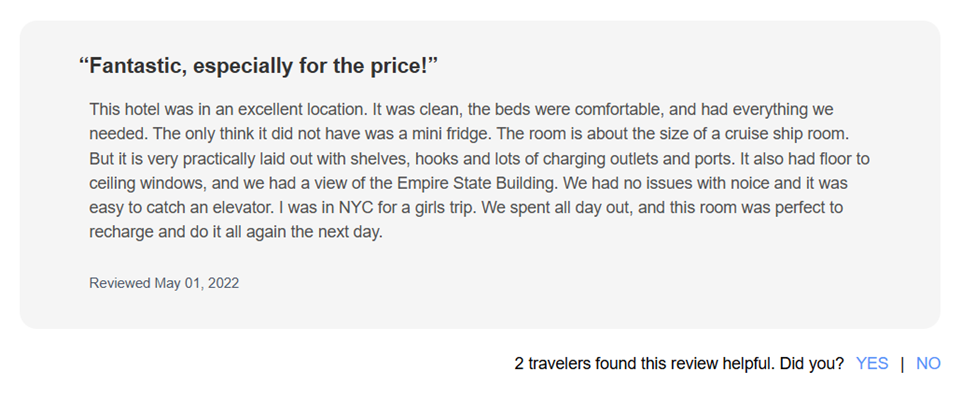}
        \caption{User review text example \label{fig:review_eg}}
\end{figure*}

To collect original human text, we gathered user reviews from an online accommodation booking platform using a Selenium-based web scraping approach. Figure \ref{fig:review_eg} illustrates an example of a review from the website.
Focusing on Manhattan, New York, we aimed to compile all available reviews for accommodations in the area during the data collection period from December 6, 2023, to January 17, 2024. The resulting dataset comprises 446 hotels in Manhattan and a total of 32,084 reviews, which include information about the hotels and the users in addition to the review text. 
Each review contains text content, rating, reviewer information, and categorical scores for the respective hotel.
In this study, we focused on text content, and additional information was used during the exploratory data analysis phase. 

To broaden our study, we utilized three additional datasets as sources of original human text: CNN news article sentences, SQuAD2 sentences, and Quora questions. These datasets are provided in \textcite{chatgpt_paraphrases_dataset}. From the approximately 420,000 entries in these datasets, we randomly sampled 10\%, yielding 8,008 CNN news samples, 9,198 SQuAD2 samples, and 24,714 Quora samples. 
All these samples are single sentences and are therefore generally shorter than our collected data. Specifically, the CNN dataset comprises sentences from news articles, while both the SQuAD2 and Quora datasets consist of single-sentence questions.

\subsection{Generating Texts \texorpdfstring{$\mathcal{G}, \mathcal{G}'$, and $\mathcal{S}$}{(G, G', and S) via LLM}}

The data points in the LLM-generated datasets ($\mathcal{G}, \mathcal{G}'$, and $\mathcal{S}$) are created by prompting the GPT-3.5-Turbo model to paraphrase the given input text. The resulting paraphrased texts are collected as data points of generated content. 
Specifically, each $i$-th text in $\mathcal{G}$ and $\mathcal{G}'$ represents the paraphrasing output of the LLM when the $i$-th text (the `Title' and `Text' of the actual review) from $\mathcal{O}$ is provided as an input. 
The $i$-th texts in $\mathcal{G}$ and $\mathcal{G}'$ are generated in parallel.
Similarly, the $i$-th text in $\mathcal{S}$ corresponds to the LLM's paraphrasing output based on the $i$-th text from $\mathcal{G}$.

In the LLM text generation procedure, the diversity of the generated text was controlled using the temperature parameter, a hyperparameter within the GPT-3.5 model. This study employed five temperature settings: {0.1, 0.4, 0.7, 1.0, 1.5}. A lower temperature yields more consistent and predictable outputs, while a higher value produces more creative and varied results. To denote the LLM-generated data, the temperature parameter used during generation is indicated by a subscript. For example, data generated with a temperature of 0.7 is represented as $\mathcal{G}_{0.7}$.

\subsection{Data Processing}

For our analysis, we transform each text in $\mathcal{O}$, $\mathcal{G}$, $\mathcal{G}'$, and $\mathcal{S}$ into a quantitative form using embeddings, where each text is mapped to a 1536-dimensional unit vector (i.e., with an $\ell_2$ norm of 1). We use the \texttt{text-embedding-3-small} model\footnote{The guidelines can be found here: https://platform.openai.com/docs/guides/embeddings} provided by OpenAI to generate these embeddings. 
This model has demonstrated good performance in clustering tasks \parencite{li2025consumer}, which aligns with the need for clustering embeddings in our analysis.
During the embedding process, texts are divided into chunks and processed chunk by chunk using the \texttt{text-embedding-3-small} model. As a pretrained model, it produces consistent output vectors that are unaffected by the chunk structure.\footnote{ To verify the robustness of the embeddings, we processed the texts twice under different conditions, and the results were identical in both cases.} 
For practical purposes, we then reduce the 1536-dimensional embedded vector to $p$ dimensions using principal component analysis.  Consequently, in our analysis, each text will be represented as a $p$-dimensional vector.

\begin{figure*}[t]
    \centering

    \begin{minipage}{\textwidth}
        \centering
        \includegraphics[width=\textwidth]{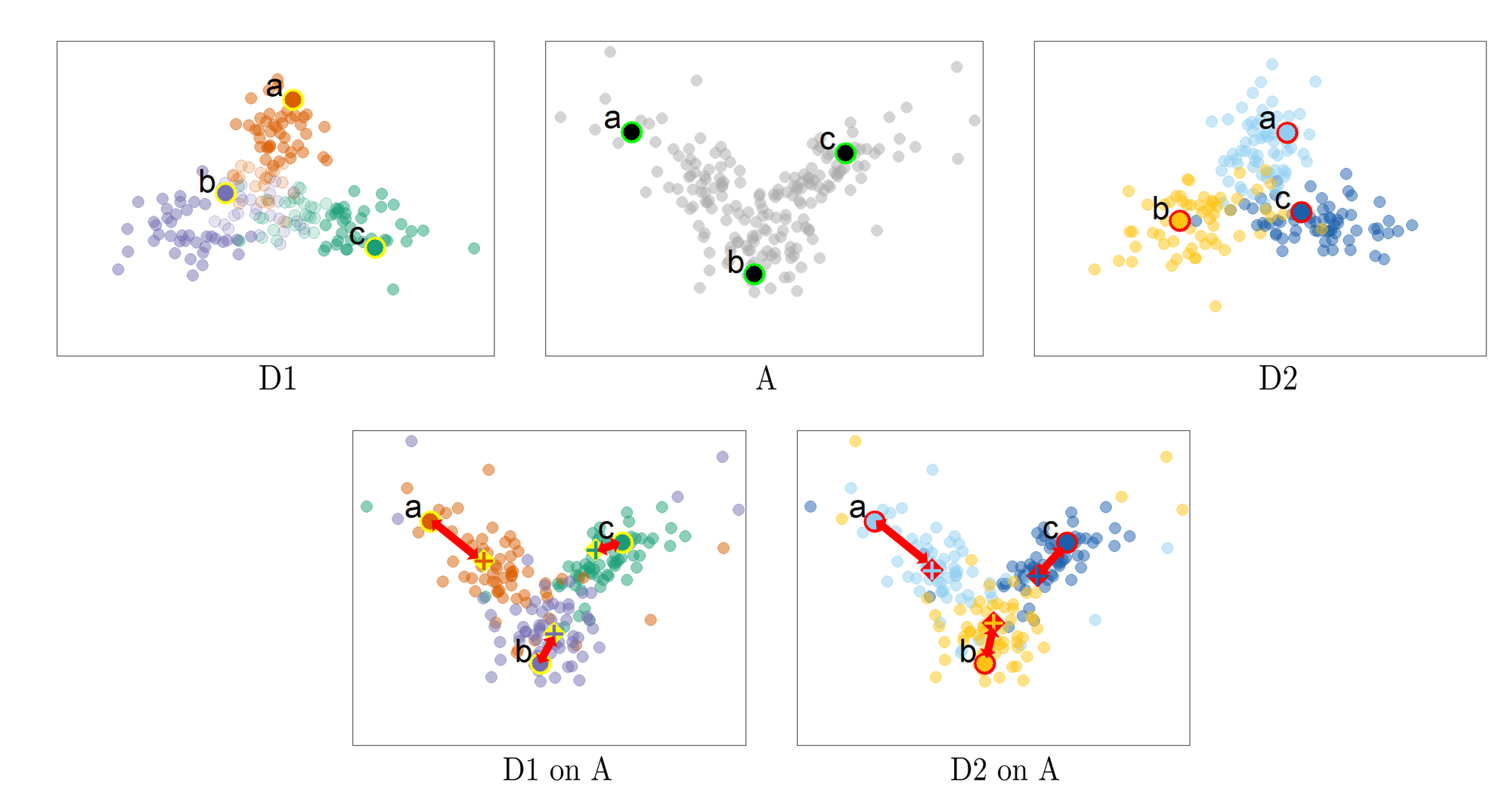}
        \captionof{figure}{An example under which the latent community structure of \(\mathbf{D}_1\) and \(\mathbf{D}_2\) are identical. 
        The points labeled \texttt{a}, \texttt{b}, and \texttt{c} appearing in all plots indicate the paired points. Across all plots, clusters are color-coded.  
        Top row: the original forms of \(\mathbf{A}\), \(\mathbf{D}_1\), and \(\mathbf{D}_2\). Bottom row: the color coding for the communities of \(\mathbf{D}_1\) and \(\mathbf{D}_2\) mapped onto \(\mathbf{A}\) from left to right, along with the corresponding community centers, marked with crosses.}
        \label{fig:null}
    \end{minipage}
    
    \vspace{1em}
    
    \begin{minipage}{\textwidth}
        \centering
        \includegraphics[width=\textwidth]{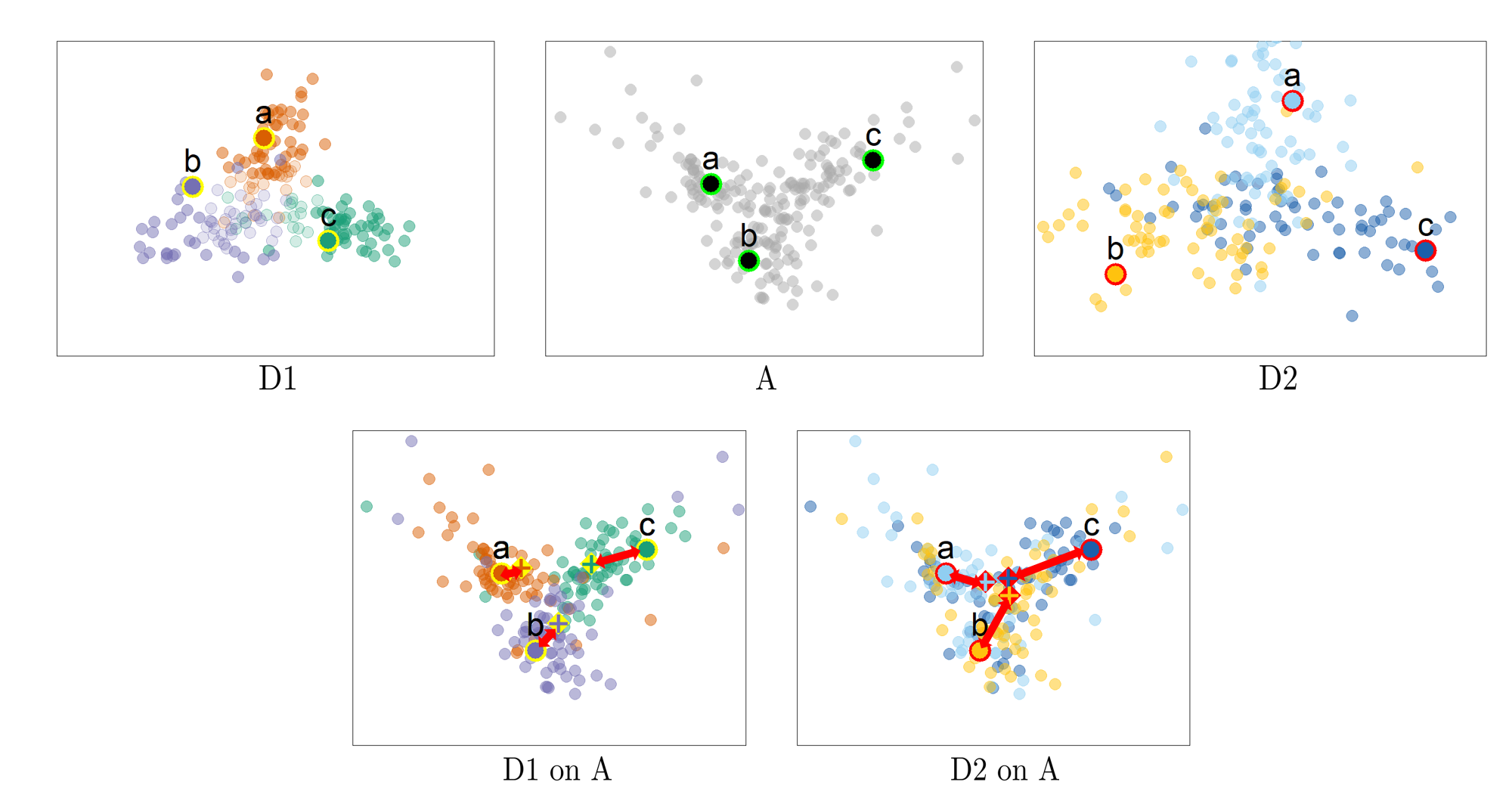}
        \captionof{figure}{An example under which the latent community structure of \(\mathbf{D}_1\) and \(\mathbf{D}_2\) are distinct. 
        The points labeled \texttt{a}, and \texttt{b}, and \texttt{c} appearing in all plots indicate the paired points. Across all plots, clusters are color-coded.  
        Top row: the original forms of \(\mathbf{A}\), \(\mathbf{D}_1\), and \(\mathbf{D}_2\). Bottom row: the color coding for \(\mathbf{D}_1\) and \(\mathbf{D}_2\) mapped onto \(\mathbf{A}\) from left to right, along with the corresponding community centers, marked with crosses.}
        \label{fig:alt}
    \end{minipage}
\end{figure*}

\section{Hypothesis Testing \label{sec:hyp}}
\subsection{Problem Setting \label{subsec:prob}}

To begin, we introduce the necessary notation. The three datasets involved in the hypothesis testing are denoted by \(\mathbf{A}\), \(\mathbf{D}_1\), and \(\mathbf{D}_2\), each of size \( n \), with their \( i \)th points paired for all \( i = 1, \ldots, n \). The dimension of \(\mathbf{A}\), \(\mathbf{D}_1\), and \(\mathbf{D}_2\) are $n \times p_a$, $n \times p_1$, and $n \times p_2$, respectively.
The objective of our hypothesis testing is to examine whether the latent community structures of the two datasets, \(\mathbf{D}_1\) and \(\mathbf{D}_2\), are identical. Denoting the community set of \(\mathbf{D}_j\) as \(\{\mathcal{C}^{(j)}_1, \ldots, \mathcal{C}^{(j)}_K\}\), where \(\mathcal{C}^{(j)}_k\) represents the index set corresponding to the \(k\)th community of dataset \(\mathbf{D}_j\), our null hypothesis can be formulated as follows:  
\begin{eqnarray}
\label{hyp:motiv_or}
\label{hyp:motiv}
    H_{0}: \{\mathcal{C}^{(1)}_1, \cdots, \mathcal{C}^{(1)}_K\} =\{\mathcal{C}^{(2)}_1, \cdots, \mathcal{C}^{(2)}_K\}.
\end{eqnarray}
Even though \(\mathbf{D}_1\) and \(\mathbf{D}_2\) may exist in different space--for instance their dimensions, \(p_1\) and \(p_2\) may be unequal--\(\mathbf{D}_1\) and \(\mathbf{D}_2\) can still share the same community structures, as the community set only involves indices. However, when \(\mathbf{D}_1\) and \(\mathbf{D}_2\) are in different spaces, directly comparing their community structures becomes challenging.

To address this, we utilize an anchor dataset \(\mathbf{A}\), enabling \(\mathbf{D}_1\) and \(\mathbf{D}_2\) to be mapped onto \(\mathbf{A}\), allowing for a quantitative comparison within the same space.  
This leads to the hypothesis introduced in Section \ref{sec:overview}:  
\begin{eqnarray}
\label{hyp:motiv}
    H_{0}(\mathbf{A}, \{\mathbf{D}_1, \mathbf{D}_2 \}) : 
    \begin{aligned}[t]
    \text{The gap in community structure between } (\mathbf{A}, \mathbf{D}_1) \\
    \text{and that between } (\mathbf{A}, \mathbf{D}_2) \text{ are the same.} 
    \end{aligned}
\end{eqnarray}
In essence, the hypothesis in \eqref{hyp:motiv} aligns with the hypothesis in \eqref{hyp:motiv_or}, which tests whether the latent community structures of \(\mathbf{D}_1\) and \(\mathbf{D}_2\) are identical.  
In the hypothesis \eqref{hyp:motiv}, the ``gap'' between \((\mathbf{A}, \mathbf{D})\) can be formalized as a statistic reflecting the mapped community structure of \(\mathbf{D}\) onto \(\mathbf{A}\). 
If \(\mathbf{D}_1\) and \(\mathbf{D}_2\) share the same community structures, their mappings onto \(\mathbf{A}\) will also be identical, leading to the acceptance of the hypothesis in \eqref{hyp:motiv}. In other words, under the null scenario of the hypothesis in \eqref{hyp:motiv_or}, the hypothesis in \eqref{hyp:motiv} will also hold. Conversely, if \(\mathbf{D}_1\) and \( \mathbf{D}_2\) have distinct community structures, their mappings onto \(\mathbf{A}\) are likely to differ, resulting in the rejection of the hypothesis in \eqref{hyp:motiv}. This corresponds to the alternative scenario of the hypothesis in \eqref{hyp:motiv_or}.

We denote the K-means clustering result on \(\mathbf{D}_j\) by \(\{C^{(j)}_{1}, \ldots, C^{(j)}_{K}\}\), where $C_k^{(j)}$ represents the index set belonging to the \( k \)th cluster of the clustering result for dataset \(\mathbf{D}_j\).
Next, denoting the \(i\)-th point of \(\mathbf{A}\) as \(A_i\), we define the center of \(C^{(j)}_{k}\) mapped onto \(\mathbf{A}\) as \(\mathbf{c}^{(j)}_k\):
\[
\mathbf{c}^{(j)}_k = \frac{1}{ \left|C^{(j)}_k\right|} \sum_{i \in C^{(j)}_k} A_i.
\]
Specifically, \(\mathbf{c}^{(j)}_k\) is the average of the data points in \(\mathbf{A}\) whose indices belong to \(C^{(j)}_k\). Additionally, we define \(d^{(j)}_i\) as the distance from \(A_i\) to its corresponding center based on the clustering result of \(\mathbf{D}_j\):
\[
d^{(j)}_i = \left\| A_i - \sum_{k=1}^{K} I\{i \in C^{(j)}_k\} \mathbf{c}^{(j)}_k \right\|_2.
\]
The set of \(d^{(j)}_i\) values is denoted by \(\mathbb{D}_{j}\) so that \(\mathbb{D}_{j} = \{d^{(j)}_1, \ldots, d^{(j)}_n\}\). 
Note that \(\mathbb{D}_{j}\) can be viewed as a set of mapped distances within clusters of \(\mathbf{D}_j\), containing information about the quantified "gap" between \(\mathbf{A}\) and \(\mathbf{D}_j\). Therefore, we examine the distributions of \(\mathbb{D}_1\) and \(\mathbb{D}_2\) to test the hypothesis in \eqref{hyp:motiv}.

We present a motivational example that illustrates the underlying concept of utilizing the distributions of \(\mathbb{D}_1\) and \(\mathbb{D}_2\) in our proposed testing procedure, as shown in Figures \ref{fig:null} and \ref{fig:alt}.
Figure \ref{fig:null} illustrates the case where the latent community structures of \(\mathbf{D}_1\) and \(\mathbf{D}_2\) are the same, while Figure \ref{fig:alt} depicts the case where the latent community structures are different.
In both plots, communities are color-coded, and the top row displays the original forms of \(\mathbf{A}\), \(\mathbf{D}_1\), and \(\mathbf{D}_2\). 
The points labeled \texttt{a}, \texttt{b}, and \texttt{c} appear in all plots to indicate the paired points. 
The bottom row shows the color coding of the communities for \(\mathbf{D}_1\) and \(\mathbf{D}_2\) mapped onto \(\mathbf{A}\), along with the corresponding mapped community centers, marked with crosses.
The plots of \(\mathbf{D}_1\) on \(\mathbf{A}\) and \(\mathbf{D}_2\) on \(\mathbf{A}\) in Figure \ref{fig:null} show that the distances from points \texttt{a},\texttt{b}, and \texttt{c}  to their corresponding community center are similar across both plots. i.e., $d_a^{(1)} \approx d_a^{(2)}, \ d_b^{(1)} \approx d_b^{(2)},  \ d_c^{(1)} \approx d_c^{(2)}$ where $d_a^{(j)}, \ d_b^{(j)}, \ d_{c}^{(j)} \in \mathbb{D}_j$.
This consistency occurs for all points and supports the conclusion of the identical distribution between \(\mathbb{D}_1\) and \(\mathbb{D}_2\). 
In contrast, the plots of \(\mathbf{D}_1\) on \(\mathbf{A}\) and \(\mathbf{D}_2\) on \(\mathbf{A}\) in Figure \ref{fig:alt} reveal a large discrepancy in the distance from those points to their corresponding community centers ($d_a^{(1)} < d_a^{(2)}, \ d_b^{(1)} < d_b^{(2)}, d_c^{(1)} < d_c^{(2)} $), a pattern likely observed across many other data points. This indicates that the distributions between \(\mathbb{D}_1\) and \(\mathbb{D}_2\) are further apart.

\subsection{Hypothesis Testing \label{subsec:hyp_test}}

In Section \ref{subsec:prob}, we discussed the idea of utilizing the distributional difference between \(\mathbb{D}_1\) and \(\mathbb{D}_2\). Specifically, we apply Johnson's paired modified t-test \parencite{johnson1978modified} to \((\mathbb{D}_1, \mathbb{D}_2)\) under permutation test framework to test the following hypothesis:  
\begin{eqnarray}
\label{hyp:med}
        \mathcal{H}_{0}( \mathbf{A}, \{\mathbb{D}_1, \mathbb{D}_2 \}) &:&  \mathbb{E} \left[\mathbf{d}_i \right] = 0, \\  
        \mathcal{H}_{1}( \mathbf{A}, \{\mathbb{D}_1, \mathbb{D}_2 \}) &:&  \mathbb{E} \left[\mathbf{d}_i \right] \neq 0 \nonumber,  
\end{eqnarray}  
where \(\mathbf{d}_i\) represents the difference between the \(i\)th paired observations in \(\mathbb{D}_1\) and \(\mathbb{D}_2\) (i.e., \(\mathbf{d}_i = d^{(1)}_i - d^{(2)}_i\)). The reported p-value corresponds to the hypothesis stated in \eqref{hyp:med}.  
To distinguish this hypothesis from earlier ones, we use the notation \(\mathcal{H}\), noting the difference in font style. Additionally, the hypothesis is written as \(\mathcal{H}_{0}( \mathbf{A}, \{\mathbb{D}_1, \mathbb{D}_2 \})\) to note that the anchor dataset is \(\mathbf{A}\), while the datasets used in the analysis are \(\mathbb{D}_j\).  

Our hypothesis procedure calculates Johnson's paired modified t-statistic, which adjusts for skewness in the dataset to accommodate the fact that \(\mathbf{d}_i\) values are non-normal with an unknown distribution. The test statistic is calculated by :
\begin{align}
\label{test_stat}
    T_{\text{mod}}
= \frac{\bar{\mathbf{d}}}{\sqrt{\hat{\sigma}^2/n}}
+ \frac{
    \hat{\mu}_{3} \cdot
    \left(
      \frac{\left(\bar{\mathbf{d}} / {\hat{\sigma}^2}\right)^2}{3}
      + \frac{1}{6 \,\hat{\sigma}^2 \,n}
    \right)
  }{
    \sqrt{\hat{\sigma}^2/n}
  }
\end{align}
where $\bar{\mathbf{d}}$ and $\hat{\sigma}$ represent the sample mean and the sample variance of $\{\mathbf{d}_1, \cdots, \mathbf{d}_n\}$, respectively, and $\hat{\mu}_{3}$ denotes the sample skewness $\hat{\mu}_{3}  = \frac{1}{n-1} \sum_{i=1}^{n} (\mathbf{d}_i - \bar{\mathbf{d}})^3$.
The p-value is calculated by referencing the test statistic generated under the null distribution, using the permutation test framework to achieve robust results, as suggested in \textcite{helwig2019robust}. The procedure for its calculation is as follows:

\begin{description}
    \item[Step 1. Calculate the sample statistic:]\hfill \\
    From the sample $\{\mathbf{d}_1, \cdots, \mathbf{d}_n\}$, calculate the test statistic in \eqref{test_stat}, and denote it by $T_{\text{mod}}$.
    \item[Step 2. Generate the samples under the null distribution using permutation test:]\hfill \\
    For \( R \) iterations, generate a new permutation sample \( \mathbf{D}^{(r)} \), where \( \mathbf{D}^{(r)} = \{ S_1^{(r)} \mathbf{d}_1, \ldots, S_n^{(r)} \mathbf{d}_n \} \) where \( S_i^{(r)} \) are  binary random variables randomly and independently drawn from \(\{-1, 1\}\).  
    \item[Step 3. Calculate the test statistic under the null distribution:]\hfill \\
    For each permutation sample \( \mathbf{D}^{(r)} \), calculate the test statistic defined in \eqref{test_stat} and denote it by \( T^{(r)} \). \item[Step 4. Compare test statistics:] Calculate the proportion of permutations where the absolute test statistic from the observed data exceeds that of the permutation samples, i.e., \[ \frac{\left|\{ T^{(r)}: \left|T_{\text{mod}}\right| > |T^{(r)}|, \ r=1, \cdots, R\}\right|}{R}. \]  
    \item[Step 5. Return the p-value:]\hfill \\
    The proportion computed in Step 4 is returned as the p-value.  
\end{description}

The overall hypothesis testing procedure is presented in Algorithm \ref{alg:tes}.

\begin{algorithm}[h]
\caption{Hypothesis Testing Procedure \label{alg:tes}}
\begin{algorithmic}[1]
    \State \textbf{Input:} Non-anchor datasets $\mathbf{D}_1$, $\mathbf{D}_2$, anchor dataset $\mathbf{A}$, the number of clusters $K$ \\
    \State $\{C^{(j)}_1, \cdots, C^{(j)}_K\} \leftarrow$ data partition of $\mathbf{D}^{(j)}$ obtained from K-means clustering for $j=1, 2$
    \For{$k = 1, \cdots, K$}
        \State $\mathbf{c}_k \leftarrow \frac{1}{|C^{(j)}_k|} \sum_{i \in C^{(j)}_k} A_i$
    \EndFor
    \For{$i = 1, \cdots, n$}
    \State $d^{(j)}_i \leftarrow \left\| A_i - \sum_{k=1}^{K} I\{i \in C^{(j)}_k\} \mathbf{c}^{(j)}_k \right\|_2$ for $j=1, 2$
    \EndFor
    \State $\mathbb{D}^{(j)} \leftarrow \{d_1^{(j)}, \cdots, d_n^{(j)}\}$ for $j=1,2$
    \State $p \leftarrow$ p-value for testing $\mathcal{H}_{0}( \mathbf{A}, \{\mathbb{D}_1, \mathbb{D}_2 \}) :  \mathbb{E} \left[\mathbf{d}_i \right] = 0$ using a permutation test \\
    \State \textbf{Output:} $p$
\end{algorithmic}
\end{algorithm}

\section{Numerical Results}\label{sec:res}

In this section, we present the results of the data analysis, focusing on the two questions discussed in Section \ref{sec:overview}:
\begin{enumerate}
    \item[Q1.] Are the latent community structures of $\mathcal{O}$ and $\mathcal{G}$ (and, equivalently, $\mathcal{G'}$) identical?
    \item[Q2.] Does the latent community structure of $\mathcal{G}$ become similar to that of $\mathcal{O}$ as the LLM parameters controlling text variability are adjusted?
\end{enumerate}
We addressed these questions under distinct parameter settings, using five temperature values of 0.1, 0.4, 0.7, 1.0, and 1.5, along with a range of cluster numbers from 2 to 5.
We present four methods for comparing two distributions: Hotelling's T-square test, Nploc test, Energytest, and Balltest. All of these tests compare the distributions of two multivariate datasets. Hotelling's \( T^2 \) test \parencite{hotelling1931generalization} and the Nploc test \parencite{helwig2019robust} are specifically designed to compare two paired multivariate datasets, testing whether the mean of the differences between the paired data is equal to a specified value.  
The Nploc test was performed using the R package \texttt{nptset} \parencite{helwig2019nptest}, which provides options to compare either the mean or the median. In this manuscript, we chose to compare the mean for consistency with our proposed method.
In contrast, Energytest \parencite{szekely2004testing} and Balltest \parencite{pan2018ball} test whether two unpaired datasets follow the same distribution.

\subsection{Analysis on Question 1 \label{subsec:hypone}}

As discussed in Section \ref{subsec:hypq1}, we tested the results for $ {H}_{0} (\mathcal{G}, \{\mathcal{O}, \mathcal{S}\})$, ${H}_{0}  (\mathcal{G}, \{\mathcal{O}, \mathcal{G}'\})$, $ {H}_{0} (\mathcal{O}, \{\mathcal{G}, \mathcal{S}\})$, and $ {H}_{0}  (\mathcal{O}, \{\mathcal{G}, \mathcal{G'}\})$ and presented the results in Table \ref{tab:hypo1}, \ref{tab:hypo2}, \ref{tab:hypo3}, and \ref{tab:hypo4}, respectively, where the null $ {H}_{0} (\mathcal{O}, \{\mathcal{G}, \mathcal{S}\})$ is expected to be rejected and the null
 \(H_{0} (\mathcal{O}, \{\mathcal{G}, \mathcal{G'}\})\) is expected to be accepted.

Table \ref{tab:hypo3} exhibits that the null hypothesis $ {H}_{0} (\mathcal{O}, \{\mathcal{G}, \mathcal{S}\})$ is rejected across all test methods and parameter settings, demonstrating that all the tested methods exhibit strong power under the alternative hypothesis scenario.
Table \ref{tab:hypo4} shows that for \(H_{0} (\mathcal{O}, \{\mathcal{G}, \mathcal{G'}\})\),  our proposed testing method accepts the null hypothesis for most of the cases. In contrast, the compared methods reject the null hypothesis in most cases, indicating that they tend to reject the null more easily.

Tables \ref{tab:hypo1} and \ref{tab:hypo2} are designed to compare human-authored text with LLM-generated text by testing $ {H}_{0} (\mathcal{G}, \{\mathcal{O}, \mathcal{S}\})$ and ${H}_{0}  (\mathcal{G}, \{\mathcal{O}, \mathcal{G}'\})$, respectively, and the test results show that the null hypothesis is rejected in most cases. There were a few exceptions, with the CNN and SQuAD2 data accounting for a large portion of these exceptions. Considering that both datasets consist of one-sentence texts and have a specific article style rather than everyday language, it is possible that LLM-generated text exhibits that kind of rigid style unless explicitly avoided. It is also observed that more rejections are exhibited for ${H}_{0}  (\mathcal{G}, \{\mathcal{O}, \mathcal{G}'\})$ in Table \ref{tab:hypo2} than in $ {H}_{0} (\mathcal{G}, \{\mathcal{O}, \mathcal{S}\})$ in Table \ref{tab:hypo1}. Considering that both hypotheses are anchored on \(\mathcal{G}\) having \(\mathcal{O}\) as one of the non-anchors, it suggests that the pair \((\mathcal{G}, \mathcal{G}')\) is more similar than the pair \((\mathcal{G}, \mathcal{S})\), implying that the effect of paraphrasing contributes more to the variation than the variation caused by the temperature parameter. One notable result is the acceptance of \(H_0 (\mathcal{G}, \{\mathcal{O}, \mathcal{S}\})\) for the Review data at \(\rho = 0.4\) and \(k = 4\) in Table \ref{tab:hypo1}, which stands out compared to the other results. The cause of this anomaly is unclear and warrants further investigation.

\begin{figure*}[t]
    \centering
    \begin{minipage}[t]{0.45\textwidth}
        \centering
        \includegraphics[width=\textwidth]{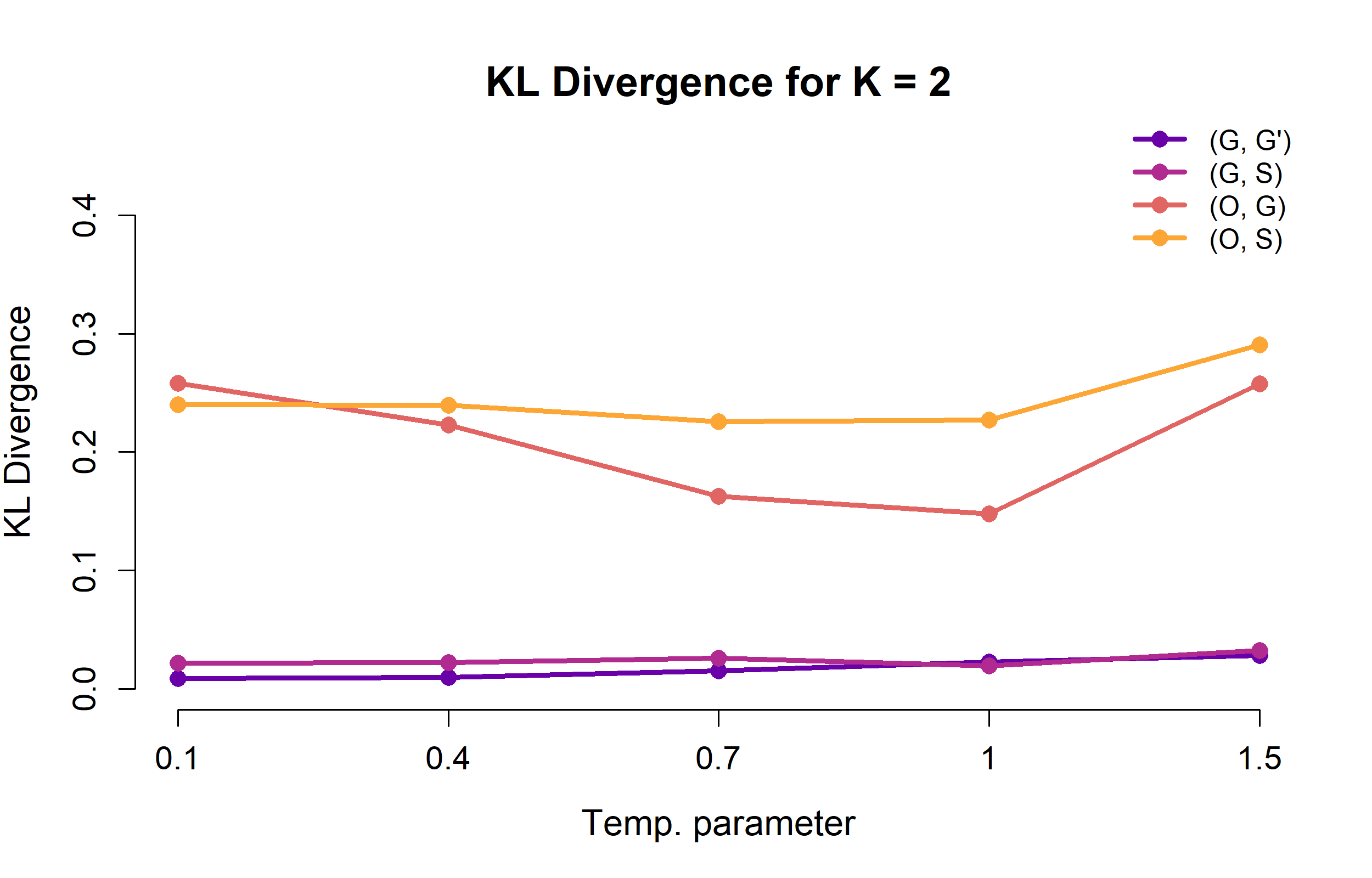}
    \end{minipage}%
    \hfill
    \begin{minipage}[t]{0.45\textwidth}
        \centering
        \includegraphics[width=\textwidth]{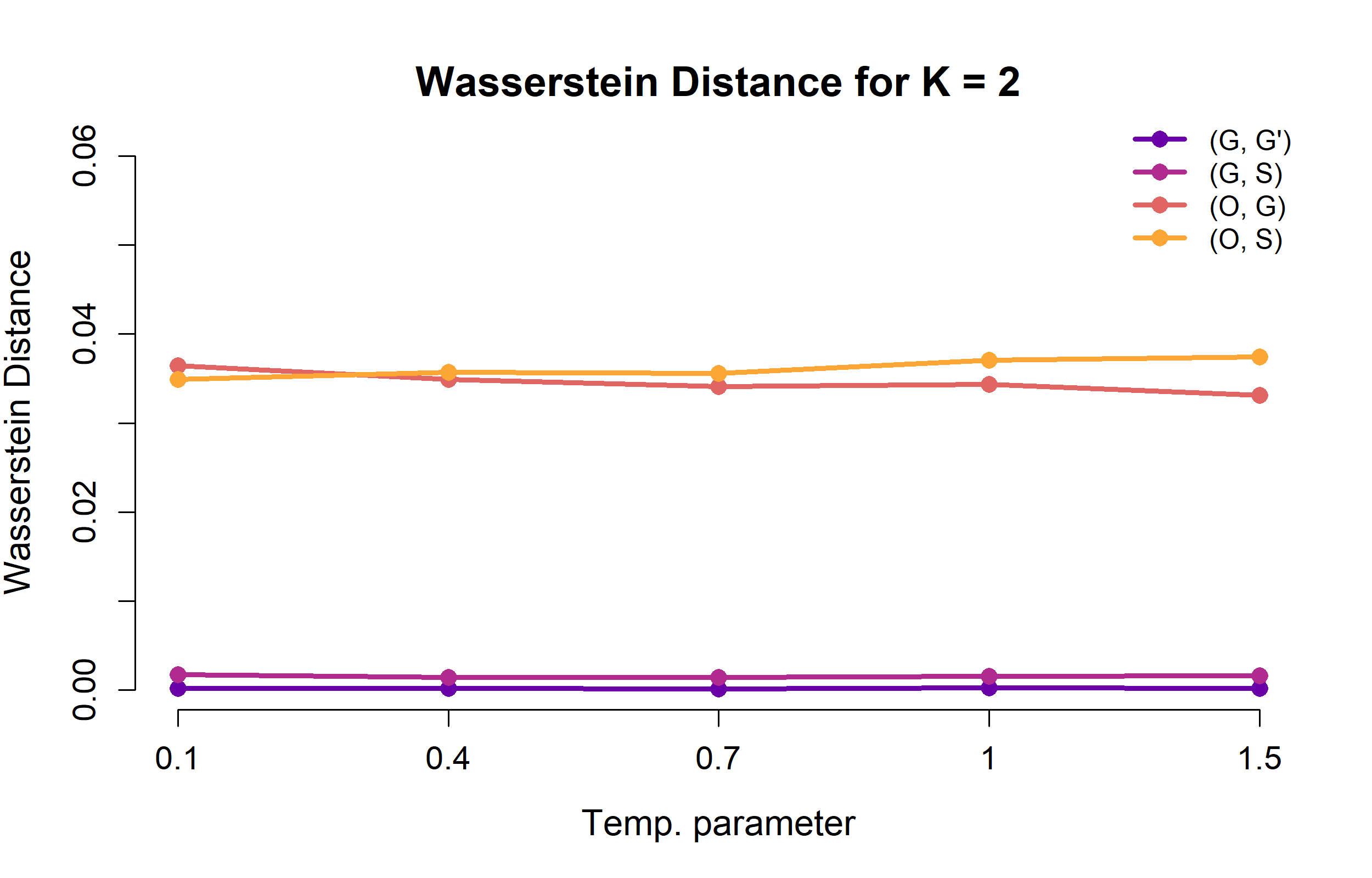}
    \end{minipage}

    \vspace{0.3cm}
    \begin{minipage}[t]{0.45\textwidth}
        \centering
        \includegraphics[width=\textwidth]{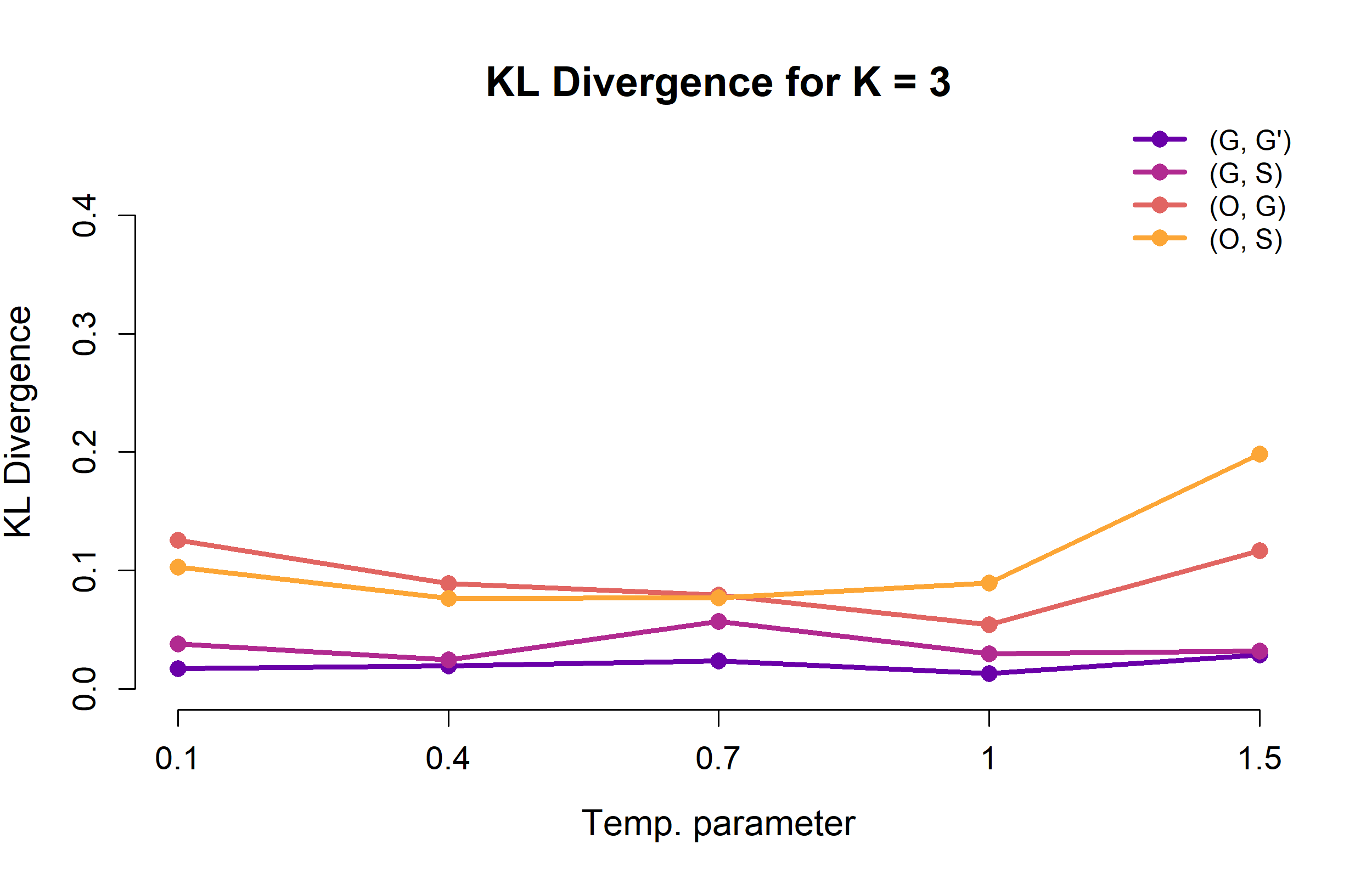}
    \end{minipage}%
    \hfill
    \begin{minipage}[t]{0.48\textwidth}
        \centering
        \includegraphics[width=\textwidth]{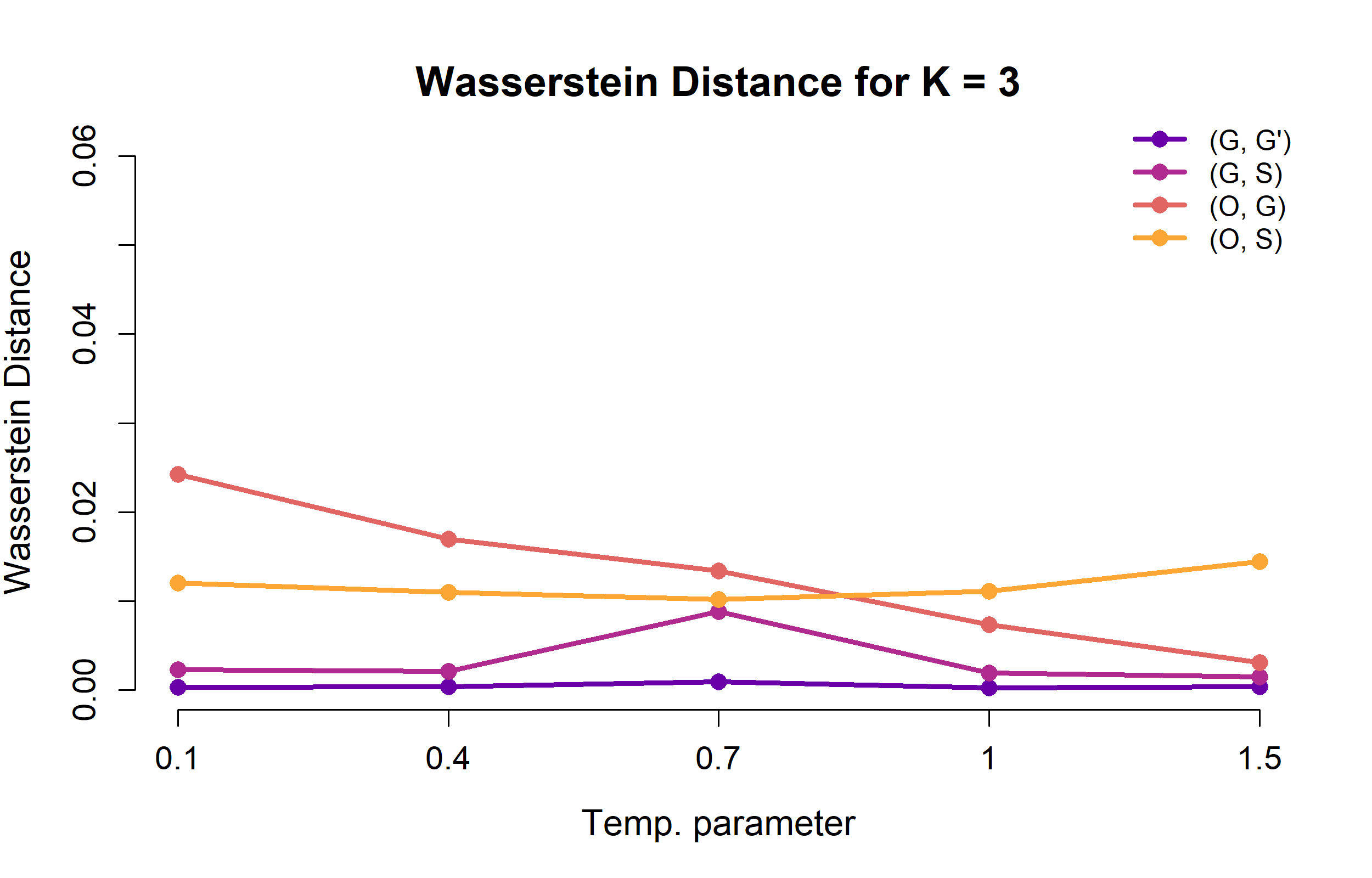}
    \end{minipage}

    \vspace{0.3cm}
    \begin{minipage}[t]{0.45\textwidth}
        \centering
        \includegraphics[width=\textwidth]{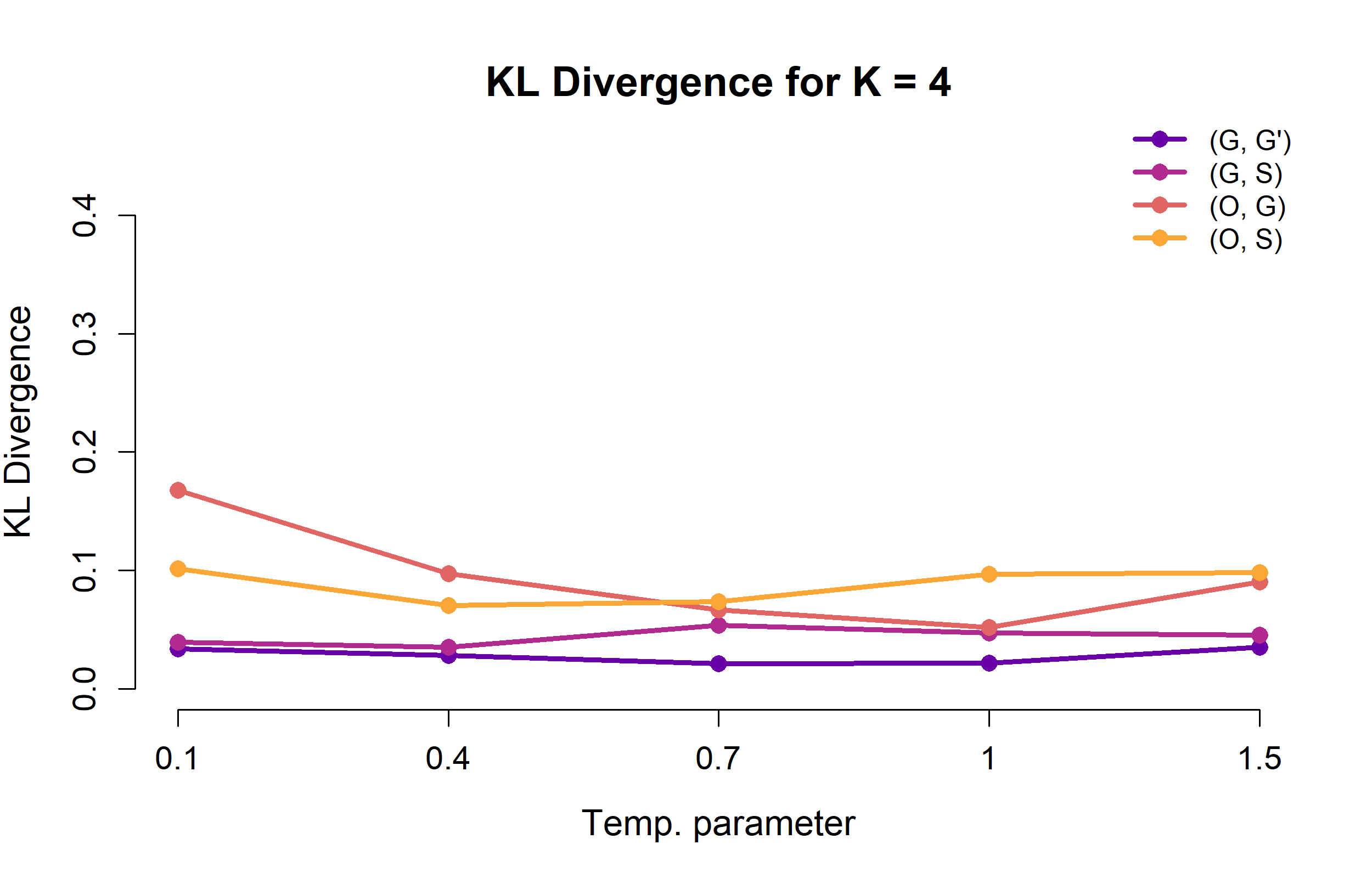}
    \end{minipage}%
    \hfill
    \begin{minipage}[t]{0.45\textwidth}
        \centering
        \includegraphics[width=\textwidth]{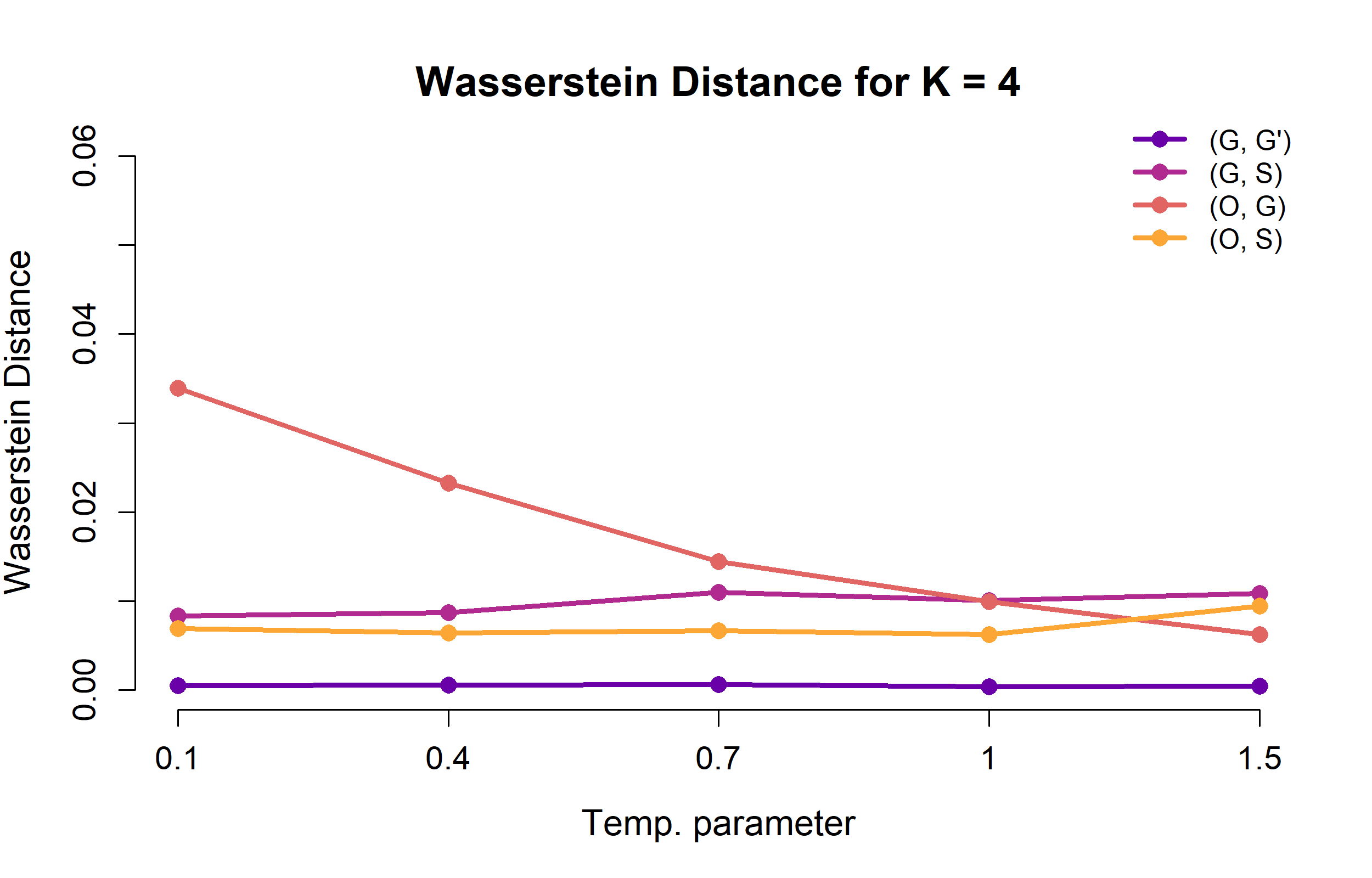}
    \end{minipage}

    \vspace{0.3cm}
    \begin{minipage}[t]{0.45\textwidth}
        \centering
        \includegraphics[width=\textwidth]{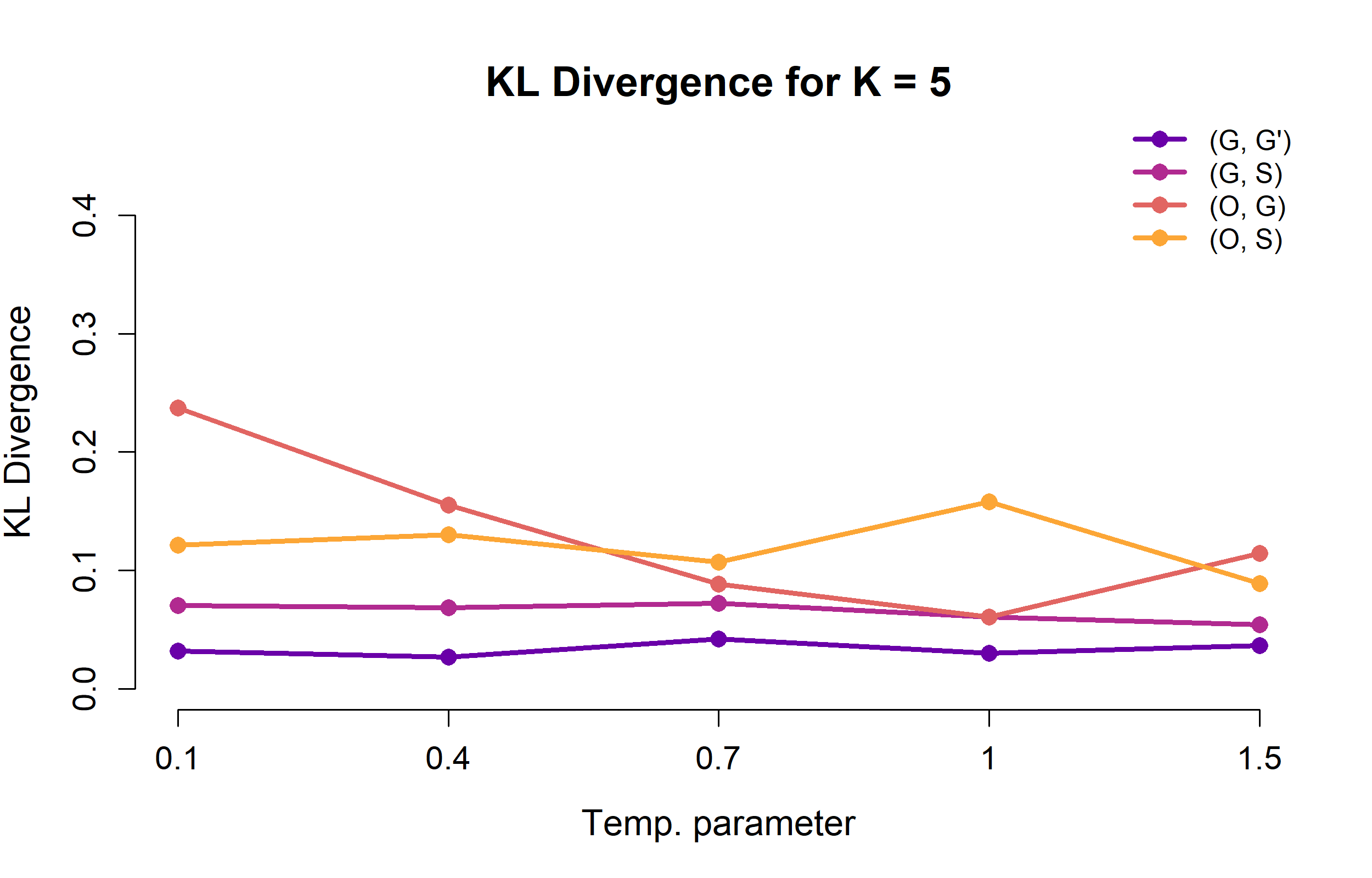}
    \end{minipage}%
    \hfill
    \begin{minipage}[t]{0.45\textwidth}
        \centering
        \includegraphics[width=\textwidth]{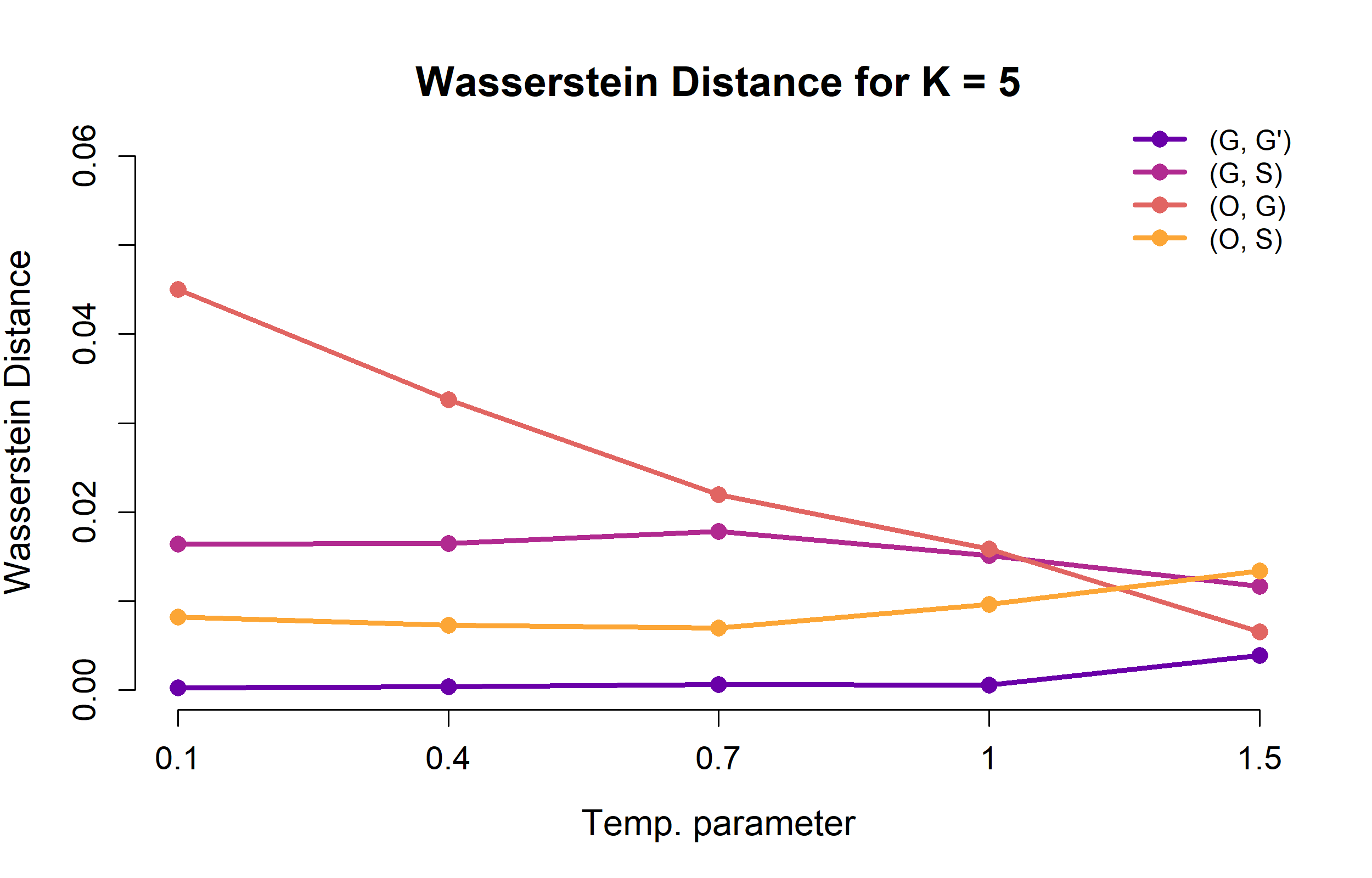}
    \end{minipage}

    \caption{Kullback-Leibler Divergence and Wasserstein Distance between \(\mathbb{D}_1\) and \(\mathbb{D}_2\) versus the temperature parameter \(\rho\) for different values of \(K\) (\(K = 2, 3, 4, 5\)) in the Review data. The settings used to calculate \(\mathbb{D}_1\) and \(\mathbb{D}_2\) in \((\mathcal{O}, \mathcal{S})\), \((\mathcal{O}, \mathcal{G'})\), \((\mathcal{G}, \mathcal{S})\), and \((\mathcal{G}, \mathcal{G'})\) correspond to those used in testing \(H_0 (\mathcal{G}, \{\mathcal{O}, \mathcal{S}\})\), \(H_0 (\mathcal{G}, \{\mathcal{O}, \mathcal{G'}\})\), \(H_0 (\mathcal{O}, \{\mathcal{G}, \mathcal{S}\})\), and \(H_0 (\mathcal{O}, \{\mathcal{G}, \mathcal{G'}\})\), respectively. Each row corresponds to a specific \( K \) value: (Top to Bottom) \( K = 2 \), \( K = 3 \), \( K = 4 \), and \( K = 5 \). Within each row, the left plot represents Kullback-Leibler  Divergence, and the right plot represents Wasserstein Distance.
}
    \label{fig:kl_ws_combined}
\end{figure*}

Additionally, we present the statistical distances between \(\mathbb{D}_1\) and \(\mathbb{D}_2\), specifically the Kullback-Leibler divergence and Wasserstein distance for the Review data. The values of \(\mathbb{D}_1\) and \(\mathbb{D}_2\) correspond to those used in the tests presented in Tables \ref{tab:hypo1}--\ref{tab:hypo4}.
The results are presented in Figure \ref{fig:kl_ws_combined} and show that the distances between \(\mathbb{D}_1\) and \(\mathbb{D}_2\) used for testing $H_0 (\mathcal{O}, \{\mathcal{G}, \ \mathcal{G}' \})$ remain consistently small across all settings. This observation aligns with the fact that the null hypothesis always holds.
On the other hand, even though the null hypothesis is expected to be rejected and is indeed rejected, the distances used to test \(H_0(\mathcal{O}, \{\mathcal{G}, \mathcal{S}\})\) remain small. This may be due to the fact that the distances were not calculated for the paired data, suggesting that, under this setting, \(\mathbb{D}_1\) and \(\mathbb{D}_2\) may be similar overall, but not when paired.
The distances between \(\mathbb{D}_1\) and \(\mathbb{D}_2\) used to test \(H_0(\mathcal{G}, \{\mathcal{O}, \mathcal{S}\})\) and \(H_0(\mathcal{G}, \{\mathcal{O}, \mathcal{G'}\})\) vary with the number of clusters \(K\), yet they remain larger than those corresponding to \(H_0(\mathcal{O}, \{\mathcal{G}, \mathcal{G'}\})\).
This suggests that investigating the gap between the distances of \(\mathbb{D}_1\) and \(\mathbb{D}_2\) corresponding to those used in testing \(H_0(\mathcal{O}, \{\mathcal{G}, \mathcal{G'}\})\) and \(H_0(\mathcal{G}, \{\mathcal{O}, \mathcal{G'}\})\) for varying values of \(K\) could aid in selecting the proper number of clusters. Since both cases involve the same three datasets ($\mathcal{O}, \mathcal{G}$, and $\mathcal{G}'$),
with the former case being constantly low, the value of \(K\) that yields a large gap may suggest that the underlying structure is well captured.
For instance, in Figure \ref{fig:kl_ws_combined}, \(K = 2\) and \(K = 3\) show a larger gap compared to \(K = 4\) and \(K = 5\), while the test for \(H_0(\mathcal{G}, \{\mathcal{O}, \mathcal{G'}\})\) was rejected in all cases. This suggests that the gap has the potential to be useful, as it does not correspond to the outcomes of the test.

\begin{table}
    \centering
    \caption{The p-values for testing the difference between $\mathcal{O}$ and $\mathcal{S}$. ``Anch.'' denotes the result from our proposed method with the anchored dataset represented in parenthesis.}
    \label{tab:hypo1}
    \belowrulesep = 0pt
    \aboverulesep = 0pt
    \small 
    \begin{tabular}{p{0.5cm} p{0.85cm} p{1.3cm} p{1.3cm} p{1.3cm} p{1.3cm} p{1.3cm} p{1.3cm} p{1.3cm} p{1.3cm}}
        \toprule
        \multirow{2}{*}{$\rho$} & \multirow{2}{*}{Dataset} & \multicolumn{8}{c}{Testing Method} \\  
        \cmidrule(lr){3-10}
        & & \multicolumn{4}{c}{Anch. ($\mathcal{G}$)} & \multirow{2}{*}{Hotelling} & \multirow{2}{*}{Nploc} & \multirow{2}{*}{Energytest} & \multirow{2}{*}{Balltest} \\            
        & & \multicolumn{1}{c}{$K=2$} & \multicolumn{1}{c}{$K=3$} & \multicolumn{1}{c}{$K=4$} & \multicolumn{1}{c}{$K=5$} & & & & \\ 
        \cmidrule(lr){1-2} \cmidrule(lr){3-6} \cmidrule(lr){7-10}
        0.1 & Review & $< 1e-3^{*}$ & $< 1e-3^{*}$ & $0.001^{*}$ & $< 1e-3^{*}$ & $< 1e-3^{*}$ & $< 1e-3^{*}$ & $0.005^{*}$ & $0.005^{*}$ \\ 
        & CNN & 0.102 & $< 1e-3^{*}$ & $< 1e-3^{*}$ & $< 1e-3^{*}$ & $< 1e-3^{*}$ & $< 1e-3^{*}$ & $0.005^{*}$ & $0.005^{*}$ \\  
        & Quora  & $< 1e-3^{*}$ & $< 1e-3^{*}$ & $< 1e-3^{*}$ & $< 1e-3^{*}$ & $< 1e-3^{*}$ & $< 1e-3^{*}$ & $0.005^{*}$ & $0.005^{*}$ \\  
        & SQuAD & $< 1e-3^{*}$ & $< 1e-3^{*}$ & $0.002^{*}$ & $< 1e-3^{*}$ & $< 1e-3^{*}$ & $< 1e-3^{*}$ & $0.005^{*}$ & $0.005^{*}$ \\  
        \cmidrule(lr){1-2} \cmidrule(lr){3-6} \cmidrule(lr){7-10}
     0.4 &  Review & $< 1e-3^{*}$ & $< 1e-3^{*}$ & $0.198$ & $< 1e-3^{*}$ & $< 1e-3^{*}$ & $< 1e-3^{*}$ & $0.005^{*}$ & $0.005^{*}$ \\      
        & CNN & 0.694 & $< 1e-3^{*}$ & $< 1e-3^{*}$ & $< 1e-3^{*}$ & $< 1e-3^{*}$ & $< 1e-3^{*}$ & $0.005^{*}$ & $0.005^{*}$ \\  
        & Quora  & $< 1e-3^{*}$ & $< 1e-3^{*}$ & $< 1e-3^{*}$ & $< 1e-3^{*}$ & $< 1e-3^{*}$ & $< 1e-3^{*}$ & $0.005^{*}$ & $0.005^{*}$ \\  
        & SQuAD & $< 1e-3^{*}$ & 0.140 & 0.938 & $< 1e-3^{*}$ & $< 1e-3^{*}$ & $< 1e-3^{*}$ & $0.005^{*}$ & $0.005^{*}$ \\      
        \cmidrule(lr){1-2} \cmidrule(lr){3-6} \cmidrule(lr){7-10}  
     0.7 & Review & $< 1e-3^{*}$ & $< 1e-3^{*}$ & $0.004^{*}$ & $< 1e-3^{*}$ & $< 1e-3^{*}$ & $< 1e-3^{*}$ & $0.005^{*}$ & $0.005^{*}$ \\      
        & CNN & $0.008^{*}$ & $< 1e-3^{*}$ & $< 1e-3^{*}$ & $< 1e-3^{*}$ & $< 1e-3^{*}$ & $< 1e-3^{*}$ & $0.005^{*}$ & $0.005^{*}$ \\  
        & Quora  & $< 1e-3^{*}$ & $< 1e-3^{*}$ & $< 1e-3^{*}$ & $< 1e-3^{*}$ & $< 1e-3^{*}$ & $< 1e-3^{*}$ & $0.005^{*}$ & $0.005^{*}$ \\  
        & SQuAD & $< 1e-3^{*}$ & $< 1e-3^{*}$ & 0.571 & $< 1e-3^{*}$ & $< 1e-3^{*}$ & $< 1e-3^{*}$ & $0.005^{*}$ & $0.005^{*}$ \\    
    \cmidrule(lr){1-2} \cmidrule(lr){3-6} \cmidrule(lr){7-10}   
     1.0 & Review & $< 1e-3^{*}$ & $< 1e-3^{*}$ & $< 1e-3^{*}$ & $< 1e-3^{*}$ & $< 1e-3^{*}$ & $< 1e-3^{*}$ & $0.005^{*}$ & $0.005^{*}$ \\      
        & CNN & 0.938 & $< 1e-3^{*}$ & $< 1e-3^{*}$ & $< 1e-3^{*}$ & $< 1e-3^{*}$ & $< 1e-3^{*}$ & $0.005^{*}$ & $0.005^{*}$ \\  
        & Quora  & $< 1e-3^{*}$ & $< 1e-3^{*}$ & $< 1e-3^{*}$ & $< 1e-3^{*}$ & $< 1e-3^{*}$ & $< 1e-3^{*}$ & $0.005^{*}$ & $0.005^{*}$ \\  
        & SQuAD & $< 1e-3^{*}$ & 0.988 & 0.171 & $< 1e-3^{*}$ & $< 1e-3^{*}$ & $< 1e-3^{*}$ & $0.005^{*}$ & $0.005^{*}$ \\  
    \cmidrule(lr){1-2} \cmidrule(lr){3-6} \cmidrule(lr){7-10} 
     1.5 & Review & $< 1e-3^{*}$ & $< 1e-3^{*}$ & $< 1e-3^{*}$ & $< 1e-3^{*}$ & $< 1e-3^{*}$ & $< 1e-3^{*}$ & $0.005^{*}$ & $0.005^{*}$ \\      
        & CNN & $< 1e-3^{*}$ & $< 1e-3^{*}$ & $< 1e-3^{*}$ & $< 1e-3^{*}$ & $< 1e-3^{*}$ & $< 1e-3^{*}$ & $0.005^{*}$ & $0.005^{*}$ \\  
        & Quora  & $< 1e-3^{*}$ & $< 1e-3^{*}$ & $< 1e-3^{*}$ & $< 1e-3^{*}$ & $< 1e-3^{*}$ & $< 1e-3^{*}$ & $0.005^{*}$ & $0.005^{*}$ \\  
        & SQuAD & $0.017^{*}$ & 0.060 & $0.006^{*}$ & $< 1e-3^{*}$ & $< 1e-3^{*}$ & $< 1e-3^{*}$ & $0.005^{*}$ & $0.005^{*}$ \\
        \bottomrule
    \end{tabular}
\end{table}

\begin{table}
    \centering
\caption{The p-values for testing the difference between $\mathcal{O}$ and $\mathcal{G}'$. ``Anch.'' denotes the result from our proposed method with the anchored dataset represented in parenthesis.}
    \label{tab:hypo2}
    \belowrulesep = 0pt
    \aboverulesep = 0pt
    \small 
    \begin{tabular}{p{0.5cm} p{0.85cm} p{1.3cm} p{1.3cm} p{1.3cm} p{1.3cm} p{1.3cm} p{1.3cm} p{1.3cm} p{1.3cm}}
        \toprule
        \multirow{2}{*}{$\rho$} & \multirow{2}{*}{Dataset} & \multicolumn{8}{c}{Testing Method} \\  
        \cmidrule(lr){3-10}
        & & \multicolumn{4}{c}{Anch. ($\mathcal{G}$)} & \multirow{2}{*}{Hotelling} & \multirow{2}{*}{Nploc} & \multirow{2}{*}{Energytest} & \multirow{2}{*}{Balltest} \\            
        & & \multicolumn{1}{c}{$K=2$} & \multicolumn{1}{c}{$K=3$} & \multicolumn{1}{c}{$K=4$} & \multicolumn{1}{c}{$K=5$} & & & & \\ 
        \cmidrule(lr){1-2} \cmidrule(lr){3-6} \cmidrule(lr){7-10} 
    0.1 & Review & $< 1e-3^{*}$ & $< 1e-3^{*}$ & $< 1e-3^{*}$ & $< 1e-3^{*}$ & $< 1e-3^{*}$ & $< 1e-3^{*}$ & $0.005^{*}$ & $0.005^{*}$ \\
       & CNN & $< 1e-3^{*}$ & $< 1e-3^{*}$ & $< 1e-3^{*}$ & $< 1e-3^{*}$ & $< 1e-3^{*}$ & $< 1e-3^{*}$ & $0.005^{*}$ & $0.005^{*}$ \\
       & Quora  & $< 1e-3^{*}$ & $< 1e-3^{*}$ & $< 1e-3^{*}$ & $< 1e-3^{*}$ & $< 1e-3^{*}$ & $< 1e-3^{*}$ & $0.005^{*}$ & $0.005^{*}$ \\
       & SQuAD & $< 1e-3^{*}$ & $< 1e-3^{*}$ & $< 1e-3^{*}$ & $< 1e-3^{*}$ & $< 1e-3^{*}$ & $< 1e-3^{*}$ & $0.005^{*}$ & $0.005^{*}$ \\
          \cmidrule(lr){1-2} \cmidrule(lr){3-6} \cmidrule(lr){7-10} 
     0.4 & Review & $< 1e-3^{*}$ & $< 1e-3^{*}$ & $< 1e-3^{*}$ & $< 1e-3^{*}$ & $< 1e-3^{*}$ & $< 1e-3^{*}$ & $0.005^{*}$ & $0.005^{*}$ \\
       & CNN & $< 1e-3^{*}$ & $< 1e-3^{*}$ & $< 1e-3^{*}$ & $< 1e-3^{*}$ & $< 1e-3^{*}$ & $< 1e-3^{*}$ & $0.005^{*}$ & $0.005^{*}$ \\
      &  Quora  & $< 1e-3^{*}$ & $< 1e-3^{*}$ & $< 1e-3^{*}$ & $< 1e-3^{*}$ & $< 1e-3^{*}$ & $< 1e-3^{*}$ & $0.005^{*}$ & $0.005^{*}$ \\
      &  SQuAD & $< 1e-3^{*}$ & $< 1e-3^{*}$ & $< 1e-3^{*}$ & $< 1e-3^{*}$ & $< 1e-3^{*}$ & $< 1e-3^{*}$ & $0.005^{*}$ & $0.005^{*}$ \\
    \cmidrule(lr){1-2} \cmidrule(lr){3-6} \cmidrule(lr){7-10}   
     0.7 & Review & $< 1e-3^{*}$ & $< 1e-3^{*}$ & $< 1e-3^{*}$ & $< 1e-3^{*}$ & $< 1e-3^{*}$ & $< 1e-3^{*}$ & $0.005^{*}$ & $0.005^{*}$ \\
       & CNN & $< 1e-3^{*}$ & $< 1e-3^{*}$ & $< 1e-3^{*}$ & $< 1e-3^{*}$ & $< 1e-3^{*}$ & $< 1e-3^{*}$ & $0.005^{*}$ & $0.005^{*}$ \\
       & Quora  & $< 1e-3^{*}$ & $< 1e-3^{*}$ & $< 1e-3^{*}$ & $< 1e-3^{*}$ & $< 1e-3^{*}$ & $< 1e-3^{*}$ & $0.005^{*}$ & $0.005^{*}$ \\
       & SQuAD & $< 1e-3^{*}$ & $< 1e-3^{*}$ & $< 1e-3^{*}$ & $< 1e-3^{*}$ & $< 1e-3^{*}$ & $< 1e-3^{*}$ & $0.005^{*}$ & $0.005^{*}$ \\
  \cmidrule(lr){1-2} \cmidrule(lr){3-6} \cmidrule(lr){7-10}   
     1.0 & Review & $< 1e-3^{*}$ & $< 1e-3^{*}$ & $< 1e-3^{*}$ & $< 1e-3^{*}$ & $< 1e-3^{*}$ & $< 1e-3^{*}$ & $0.005^{*}$ & $0.005^{*}$ \\
       & CNN & $0.030^{*}$ & $< 1e-3^{*}$ & $< 1e-3^{*}$ & $< 1e-3^{*}$ & $< 1e-3^{*}$ & $< 1e-3^{*}$ & $0.005^{*}$ & $0.005^{*}$ \\
       & Quora  & 0.113 & $< 1e-3^{*}$ & $< 1e-3^{*}$ & $< 1e-3^{*}$ & $< 1e-3^{*}$ & $< 1e-3^{*}$ & $0.005^{*}$ & $0.005^{*}$ \\
       & SQuAD & 0.964 & $< 1e-3^{*}$ & $< 1e-3^{*}$ & $< 1e-3^{*}$ & $< 1e-3^{*}$ & $< 1e-3^{*}$ & $0.005^{*}$ & $0.005^{*}$ \\
  \cmidrule(lr){1-2} \cmidrule(lr){3-6} \cmidrule(lr){7-10}   
     1.5 & Review & $< 1e-3^{*}$ & $< 1e-3^{*}$ & $< 1e-3^{*}$ & $0.024^{*}$ & $< 1e-3^{*}$ & $< 1e-3^{*}$ & $0.005^{*}$ & $0.005^{*}$ \\
       & CNN & 0.813 & $< 1e-3^{*}$ & $< 1e-3^{*}$ & $< 1e-3^{*}$ & $< 1e-3^{*}$ & $< 1e-3^{*}$ & $0.005^{*}$ & $0.005^{*}$ \\
       & Quora  & $< 1e-3^{*}$ & $< 1e-3^{*}$ & $< 1e-3^{*}$ & $< 1e-3^{*}$ & $< 1e-3^{*}$ & $< 1e-3^{*}$ & $0.005^{*}$ & $0.005^{*}$ \\
       & SQuAD & $< 1e-3^{*}$ & $< 1e-3^{*}$ & $0.002^{*}$ & $0.003^{*}$ & $< 1e-3^{*}$ & $< 1e-3^{*}$ & $0.005^{*}$ & $0.005^{*}$ \\
    \bottomrule
    \end{tabular}
\end{table}

\begin{table}
    \centering
\caption{The p-values for testing the difference between $\mathcal{G}$ and $\mathcal{S}$. ``Anch.'' denotes the result from our proposed method with the anchored dataset represented in parenthesis.}
    \label{tab:hypo3}
    \belowrulesep = 0pt
    \aboverulesep = 0pt
    \small 
    \begin{tabular}{p{0.5cm} p{0.85cm} p{1.3cm} p{1.3cm} p{1.3cm} p{1.3cm} p{1.3cm} p{1.3cm} p{1.3cm} p{1.3cm}}
        \toprule
        \multirow{2}{*}{$\rho$} & \multirow{2}{*}{Dataset} & \multicolumn{8}{c}{Testing Method} \\  
        \cmidrule(lr){3-10}
        & & \multicolumn{4}{c}{Anch. ($\mathcal{G}$)} & \multirow{2}{*}{Hotelling} & \multirow{2}{*}{Nploc} & \multirow{2}{*}{Energytest} & \multirow{2}{*}{Balltest} \\            
        & & \multicolumn{1}{c}{$K=2$} & \multicolumn{1}{c}{$K=3$} & \multicolumn{1}{c}{$K=4$} & \multicolumn{1}{c}{$K=5$} & & & & \\ 
        \cmidrule(lr){1-2} \cmidrule(lr){3-6} \cmidrule(lr){7-10} 
        0.1 & Review & $< 1e-3^{*}$ & $< 1e-3^{*}$ & $< 1e-3^{*}$ & $< 1e-3^{*}$ & $< 1e-3^{*}$ & $< 1e-3^{*}$ & $0.005^{*}$ & $0.005^{*}$ \\      
        & CNN & $< 1e-3^{*}$ & $< 1e-3^{*}$ & $< 1e-3^{*}$ & $< 1e-3^{*}$ & $< 1e-3^{*}$ & $< 1e-3^{*}$ & $0.005^{*}$ & $0.005^{*}$ \\  
        & Quora  & $< 1e-3^{*}$ & $< 1e-3^{*}$ & $< 1e-3^{*}$ & $< 1e-3^{*}$ & $< 1e-3^{*}$ & $< 1e-3^{*}$ & $0.005^{*}$ & $0.005^{*}$ \\  
        & SQuAD & $< 1e-3^{*}$ & $< 1e-3^{*}$ & $< 1e-3^{*}$ & $< 1e-3^{*}$ & $< 1e-3^{*}$ & $< 1e-3^{*}$ & $0.005^{*}$ & $0.005^{*}$ \\  
          \cmidrule(lr){1-2} \cmidrule(lr){3-6} \cmidrule(lr){7-10}
     0.4 & Review & $< 1e-3^{*}$ & $< 1e-3^{*}$ & $< 1e-3^{*}$ & $< 1e-3^{*}$ & $< 1e-3^{*}$ & $< 1e-3^{*}$ & $0.005^{*}$ & $0.005^{*}$ \\      
        & CNN & $< 1e-3^{*}$ & $< 1e-3^{*}$ & $< 1e-3^{*}$ & $< 1e-3^{*}$ & $< 1e-3^{*}$ & $< 1e-3^{*}$ & $0.005^{*}$ & $0.005^{*}$ \\  
        & Quora  & $< 1e-3^{*}$ & $< 1e-3^{*}$ & $< 1e-3^{*}$ & $< 1e-3^{*}$ & $< 1e-3^{*}$ & $< 1e-3^{*}$ & $0.005^{*}$ & $0.005^{*}$ \\  
        & SQuAD & $< 1e-3^{*}$ & $< 1e-3^{*}$ & $< 1e-3^{*}$ & $< 1e-3^{*}$ & $< 1e-3^{*}$ & $< 1e-3^{*}$ & $0.005^{*}$ & $0.005^{*}$ \\      
    \cmidrule(lr){1-2} \cmidrule(lr){3-6} \cmidrule(lr){7-10}
     0.7 & Review & $< 1e-3^{*}$ & $< 1e-3^{*}$ & $< 1e-3^{*}$ & $< 1e-3^{*}$ & $< 1e-3^{*}$ & $< 1e-3^{*}$ & $0.005^{*}$ & $0.005^{*}$ \\      
        & CNN & $< 1e-3^{*}$ & $< 1e-3^{*}$ & $< 1e-3^{*}$ & $< 1e-3^{*}$ & $< 1e-3^{*}$ & $< 1e-3^{*}$ & $0.005^{*}$ & $0.005^{*}$ \\  
        & Quora  & $< 1e-3^{*}$ & $< 1e-3^{*}$ & $< 1e-3^{*}$ & $< 1e-3^{*}$ & $< 1e-3^{*}$ & $< 1e-3^{*}$ & $0.005^{*}$ & $0.005^{*}$ \\  
        & SQuAD & $< 1e-3^{*}$ & $< 1e-3^{*}$ & $< 1e-3^{*}$ & $< 1e-3^{*}$ & $< 1e-3^{*}$ & $< 1e-3^{*}$ & $0.005^{*}$ & $0.005^{*}$ \\    
  \cmidrule(lr){1-2} \cmidrule(lr){3-6} \cmidrule(lr){7-10}
     1.0 & Review & $< 1e-3^{*}$ & $< 1e-3^{*}$ & $< 1e-3^{*}$ & $< 1e-3^{*}$ & $< 1e-3^{*}$ & $< 1e-3^{*}$ & $0.005^{*}$ & $0.005^{*}$ \\      
        & CNN & $< 1e-3^{*}$ & $< 1e-3^{*}$ & $< 1e-3^{*}$ & $< 1e-3^{*}$ & $< 1e-3^{*}$ & $< 1e-3^{*}$ & $0.005^{*}$ & $0.005^{*}$ \\  
        & Quora  & $< 1e-3^{*}$ & $< 1e-3^{*}$ & $< 1e-3^{*}$ & $< 1e-3^{*}$ & $< 1e-3^{*}$ & $< 1e-3^{*}$ & $0.005^{*}$ & $0.005^{*}$ \\  
        & SQuAD & $< 1e-3^{*}$ & $< 1e-3^{*}$ & $< 1e-3^{*}$ & $< 1e-3^{*}$ & $< 1e-3^{*}$ & $< 1e-3^{*}$ & $0.005^{*}$ & $0.005^{*}$ \\  
  \cmidrule(lr){1-2} \cmidrule(lr){3-6} \cmidrule(lr){7-10}
     1.5 & Review & $< 1e-3^{*}$ & $< 1e-3^{*}$ & $< 1e-3^{*}$ & $< 1e-3^{*}$ & $< 1e-3^{*}$ & $< 1e-3^{*}$ & $0.005^{*}$ & $0.005^{*}$ \\      
        & CNN & $< 1e-3^{*}$ & $< 1e-3^{*}$ & $< 1e-3^{*}$ & $< 1e-3^{*}$ & $< 1e-3^{*}$ & $< 1e-3^{*}$ & $0.005^{*}$ & $0.005^{*}$ \\  
        & Quora  & $< 1e-3^{*}$ & $< 1e-3^{*}$ & $< 1e-3^{*}$ & $< 1e-3^{*}$ & $< 1e-3^{*}$ & $< 1e-3^{*}$ & $0.005^{*}$ & $0.005^{*}$ \\  
        & SQuAD & $< 1e-3^{*}$ & $< 1e-3^{*}$ & $< 1e-3^{*}$ & $< 1e-3^{*}$ & $< 1e-3^{*}$ & $< 1e-3^{*}$ & $0.005^{*}$ & $0.005^{*}$ \\     
    \bottomrule
    \end{tabular}
\end{table}

\begin{table}
    \centering
\caption{The p-values for testing the difference between $\mathcal{G}$ and $\mathcal{G}'$. ``Anch.'' denotes the result from our proposed method with the anchored dataset represented in parenthesis.}
    \label{tab:hypo4}
    \belowrulesep = 0pt
    \aboverulesep = 0pt
    \small 
    \begin{tabular}{p{0.5cm} p{0.85cm} p{1.3cm} p{1.3cm} p{1.3cm} p{1.3cm} p{1.3cm} p{1.3cm} p{1.3cm} p{1.3cm}}
        \toprule
        \multirow{2}{*}{$\rho$} & \multirow{2}{*}{Dataset} & \multicolumn{8}{c}{Testing Method} \\  
        \cmidrule(lr){3-10}
        & & \multicolumn{4}{c}{Anch. ($\mathcal{G}$)} & \multirow{2}{*}{Hotelling} & \multirow{2}{*}{Nploc} & \multirow{2}{*}{Energytest} & \multirow{2}{*}{Balltest} \\            
        & & \multicolumn{1}{c}{$K=2$} & \multicolumn{1}{c}{$K=3$} & \multicolumn{1}{c}{$K=4$} & \multicolumn{1}{c}{$K=5$} & & & & \\ 
        \cmidrule(lr){1-2} \cmidrule(lr){3-6} \cmidrule(lr){7-10} 
        0.1 & Review & 0.644 & 0.154 & $0.020^{*}$ & 0.870 & $< 1e-3^{*}$ & $< 1e-3^{*}$ & $0.005^{*}$ & $0.005^{*}$ \\
        & CNN & $< 1e-3^{*}$ & 0.071 & 0.688 & 0.154 & $< 1e-3^{*}$ & $< 1e-3^{*}$ & $0.005^{*}$ & $0.005^{*}$ \\  
        & Quora  & $< 1e-3^{*}$ & $< 1e-3^{*}$ & $< 1e-3^{*}$ & $< 1e-3^{*}$ & $< 1e-3^{*}$ & $< 1e-3^{*}$ & 0.109 & 0.359 \\
        & SQuAD & $< 1e-3^{*}$ & 0.949 & $0.017^{*}$ & 0.215 & $< 1e-3^{*}$ & $< 1e-3^{*}$ & $0.005^{*}$ & $0.005^{*}$ \\
         \cmidrule(lr){1-2} \cmidrule(lr){3-6} \cmidrule(lr){7-10}  
     0.4 & Review & 0.347 & $0.039^{*}$ & 0.051 & 0.367 & $< 1e-3^{*}$ & $< 1e-3^{*}$ & $0.005^{*}$ & $0.005^{*}$ \\
        & CNN & 0.576 & 0.096 & 0.345 & 0.076 & $< 1e-3^{*}$ & $< 1e-3^{*}$ & $0.005^{*}$ & $0.005^{*}$ \\  
        & Quora  & $0.020^{*}$ & $< 1e-3^{*}$ & 0.316 & 0.158 & $< 1e-3^{*}$ & $< 1e-3^{*}$ & $0.005^{*}$ & $0.005^{*}$ \\
        & SQuAD & $0.016^{*}$ & 0.129 & 0.804 & 0.640 & $< 1e-3^{*}$ & $< 1e-3^{*}$ & $0.005^{*}$ & $0.005^{*}$ \\
    \cmidrule(lr){1-2} \cmidrule(lr){3-6} \cmidrule(lr){7-10} 
     0.7 & Review & 0.564 & $< 1e-3^{*}$ & $0.032^{*}$ & 0.117 & $< 1e-3^{*}$ & $< 1e-3^{*}$ & $0.005^{*}$ & $0.005^{*}$ \\
        & CNN & 0.056 & 0.828 & 0.323 & 0.122 & 0.974 & 0.943 & 0.996 & 0.969 \\  
        & Quora  & 0.782 & 0.355 & $0.009^{*}$ & 0.757 & $< 1e-3^{*}$ & $< 1e-3^{*}$ & $0.005^{*}$ & $0.005^{*}$ \\
        & SQuAD & 0.451 & 0.409 & 0.398 & $< 1e-3^{*}$ & 0.517 & 0.631 & 0.998 & 0.998 \\
  \cmidrule(lr){1-2} \cmidrule(lr){3-6} \cmidrule(lr){7-10} 
     1.0 & Review & 0.098 & 0.453 & 0.341 & 0.333 & $< 1e-3^{*}$ & $< 1e-3^{*}$ & $0.005^{*}$ & $0.005^{*}$ \\
        & CNN & 0.189 & $0.002^{*}$ & 0.065 & $0.010^{*}$ & 0.347 & 0.295 & 0.851 & 0.709 \\  
        & Quora  & 0.740 & 0.858 & $0.005^{*}$ & $0.011^{*}$ & 0.638 & 0.497 & 0.959 & 0.909 \\
        & SQuAD & 0.548 & 0.357 & 0.988 & 0.229 & $< 1e-3^{*}$ & $< 1e-3^{*}$ & $0.005^{*}$ & $0.005^{*}$ \\
  \cmidrule(lr){1-2} \cmidrule(lr){3-6} \cmidrule(lr){7-10} 
     1.5 & Review & 0.604 & 0.660 & 0.911 & $< 1e-3^{*}$ & $< 1e-3^{*}$ & $< 1e-3^{*}$ & $0.005^{*}$ & $0.005^{*}$ \\
        & CNN & $< 1e-3^{*}$ & 0.481 & $0.045^{*}$ & $0.026^{*}$ & $0.028^{*}$ & 0.053 & $0.016^{*}$ & $0.048^{*}$ \\  
        & Quora  & 0.352 & 0.257 & 0.227 & 0.105 & $< 1e-3^{*}$ & $< 1e-3^{*}$ & $0.005^{*}$ & $0.005^{*}$ \\
        & SQuAD & 0.073 & 0.375 & 0.848 & 0.233 & $< 1e-3^{*}$ & $< 1e-3^{*}$ & $0.005^{*}$ & $0.005^{*}$ \\
    \bottomrule
    \end{tabular}
\end{table}

\begin{table}
    \centering
\caption{The p-values for testing the difference between $\mathcal{O}$ and $\mathcal{G}_{\rho_2}$.  For our proposed method, $\mathcal{G}_{0.1}$ is used as the anchor dataset. ``Anch.'' denotes the result from our proposed method with the anchored dataset represented in parenthesis. }
    \label{tab:hypo5}
    \belowrulesep = 0pt
    \aboverulesep = 0pt
    \small 
    \begin{tabular}{p{1.1cm} p{0.85cm} p{1.3cm} p{1.3cm} p{1.3cm} p{1.3cm} p{1.3cm} p{1.3cm} p{1.3cm} p{1.3cm}}
        \toprule
        \multirow{2}{*}{$\rho$ pair} & \multirow{2}{*}{Dataset} & \multicolumn{8}{c}{Testing Method} \\  
        \cmidrule(lr){3-10}
        & & \multicolumn{4}{c}{Anch. ($\mathcal{G}$)} & \multirow{2}{*}{Hotelling} & \multirow{2}{*}{Nploc} & \multirow{2}{*}{Energytest} & \multirow{2}{*}{Balltest} \\            
        & & \multicolumn{1}{c}{$K=2$} & \multicolumn{1}{c}{$K=3$} & \multicolumn{1}{c}{$K=4$} & \multicolumn{1}{c}{$K=5$} & & & & \\ 
        \cmidrule(lr){1-2} \cmidrule(lr){3-6} \cmidrule(lr){7-10} 
        $(0.1, 0.4)$ 
        & Review & $< 1e-3^{*}$ & $< 1e-3^{*}$ & $< 1e-3^{*}$ & $< 1e-3^{*}$ & $< 1e-3^{*}$ & $< 1e-3^{*}$ & $0.005^{*}$ & $0.005^{*}$ \\
        & CNN    & $< 1e-3^{*}$ & $< 1e-3^{*}$ & $< 1e-3^{*}$ & $< 1e-3^{*}$ & $< 1e-3^{*}$ & $< 1e-3^{*}$ & $0.005^{*}$ & $0.005^{*}$ \\
        & Quora  & $< 1e-3^{*}$ & $< 1e-3^{*}$ & $< 1e-3^{*}$ & $< 1e-3^{*}$ & $< 1e-3^{*}$ & $< 1e-3^{*}$ & $0.005^{*}$ & $0.005^{*}$ \\
        & SQuAD & $< 1e-3^{*}$ & $< 1e-3^{*}$ & $< 1e-3^{*}$ & $< 1e-3^{*}$ & $< 1e-3^{*}$ & $< 1e-3^{*}$ & $0.005^{*}$ & $0.005^{*}$ \\
        \cmidrule(lr){1-2} \cmidrule(lr){3-6} \cmidrule(lr){7-10}
        $(0.1, 0.7)$ 
        & Review & $< 1e-3^{*}$ & $< 1e-3^{*}$ & $< 1e-3^{*}$ & $< 1e-3^{*}$ & $< 1e-3^{*}$ & $< 1e-3^{*}$ & $0.005^{*}$ & $0.005^{*}$ \\
        & CNN    & $< 1e-3^{*}$ & $< 1e-3^{*}$ & $< 1e-3^{*}$ & $< 1e-3^{*}$ & $< 1e-3^{*}$ & $< 1e-3^{*}$ & $0.005^{*}$ & $0.005^{*}$ \\
        & Quora  & $< 1e-3^{*}$ & $< 1e-3^{*}$ & $< 1e-3^{*}$ & $< 1e-3^{*}$ & $< 1e-3^{*}$ & $< 1e-3^{*}$ & $0.005^{*}$ & $0.005^{*}$ \\
        & SQuAD & $< 1e-3^{*}$ & $< 1e-3^{*}$ & $< 1e-3^{*}$ & $< 1e-3^{*}$ & $< 1e-3^{*}$ & $< 1e-3^{*}$ & $0.005^{*}$ & $0.005^{*}$ \\
        \cmidrule(lr){1-2} \cmidrule(lr){3-6} \cmidrule(lr){7-10}
        $(0.1, 1.0)$ 
        & Review & $< 1e-3^{*}$ & $< 1e-3^{*}$ & $< 1e-3^{*}$ & $< 1e-3^{*}$ & $< 1e-3^{*}$ & $< 1e-3^{*}$ & $0.005^{*}$ & $0.005^{*}$ \\
        & CNN    & $< 1e-3^{*}$ & $< 1e-3^{*}$ & $< 1e-3^{*}$ & $< 1e-3^{*}$ & $< 1e-3^{*}$ & $< 1e-3^{*}$ & $0.005^{*}$ & $0.005^{*}$ \\
        & Quora  & $< 1e-3^{*}$ & $< 1e-3^{*}$ & $< 1e-3^{*}$ & $< 1e-3^{*}$ & $< 1e-3^{*}$ & $< 1e-3^{*}$ & $0.005^{*}$ & $0.005^{*}$ \\
        & SQuAD & 0.095 & $< 1e-3^{*}$ & $< 1e-3^{*}$ & $< 1e-3^{*}$ & $< 1e-3^{*}$ & $< 1e-3^{*}$ & $0.005^{*}$ & $0.005^{*}$ \\
        \cmidrule(lr){1-2} \cmidrule(lr){3-6} \cmidrule(lr){7-10}
        $(0.1, 1.5)$ 
        & Review & $< 1e-3^{*}$ & $0.020^{*}$ & $< 1e-3^{*}$ & $< 1e-3^{*}$ & $< 1e-3^{*}$ & $< 1e-3^{*}$ & $0.005^{*}$ & $0.005^{*}$ \\
        & CNN    & $0.030^{*}$ & $< 1e-3^{*}$ & $< 1e-3^{*}$ & $< 1e-3^{*}$ & $< 1e-3^{*}$ & $< 1e-3^{*}$ & $0.005^{*}$ & $0.005^{*}$ \\
        & Quora  & $< 1e-3^{*}$ & $< 1e-3^{*}$ & $< 1e-3^{*}$ & $< 1e-3^{*}$ & $< 1e-3^{*}$ & $< 1e-3^{*}$ & $0.005^{*}$ & $0.005^{*}$ \\
        & SQuAD & $< 1e-3^{*}$ & $0.023^{*}$ & 0.945 & 0.256 & $< 1e-3^{*}$ & $< 1e-3^{*}$ & $0.005^{*}$ & $0.005^{*}$ \\
        \cmidrule(lr){1-2} \cmidrule(lr){3-6} \cmidrule(lr){7-10}
    \bottomrule
    \end{tabular}
\end{table}


\begin{figure*}[t]
    \centering
    \begin{minipage}[t]{0.48\textwidth}
        \centering
        \includegraphics[width=\textwidth]{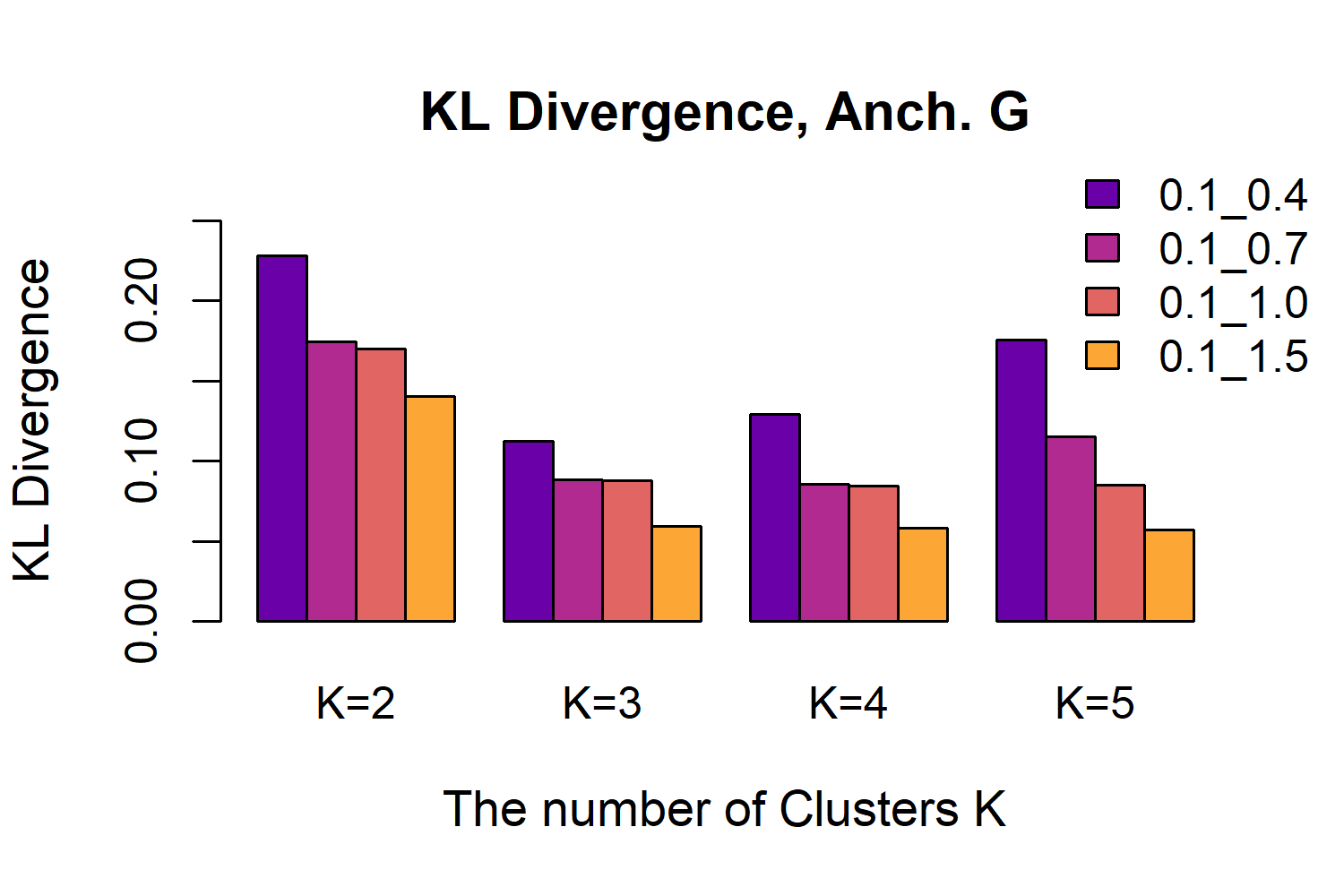}
    \end{minipage}%
    \hfill
    \begin{minipage}[t]{0.48\textwidth}
        \centering
        \includegraphics[width=\textwidth]{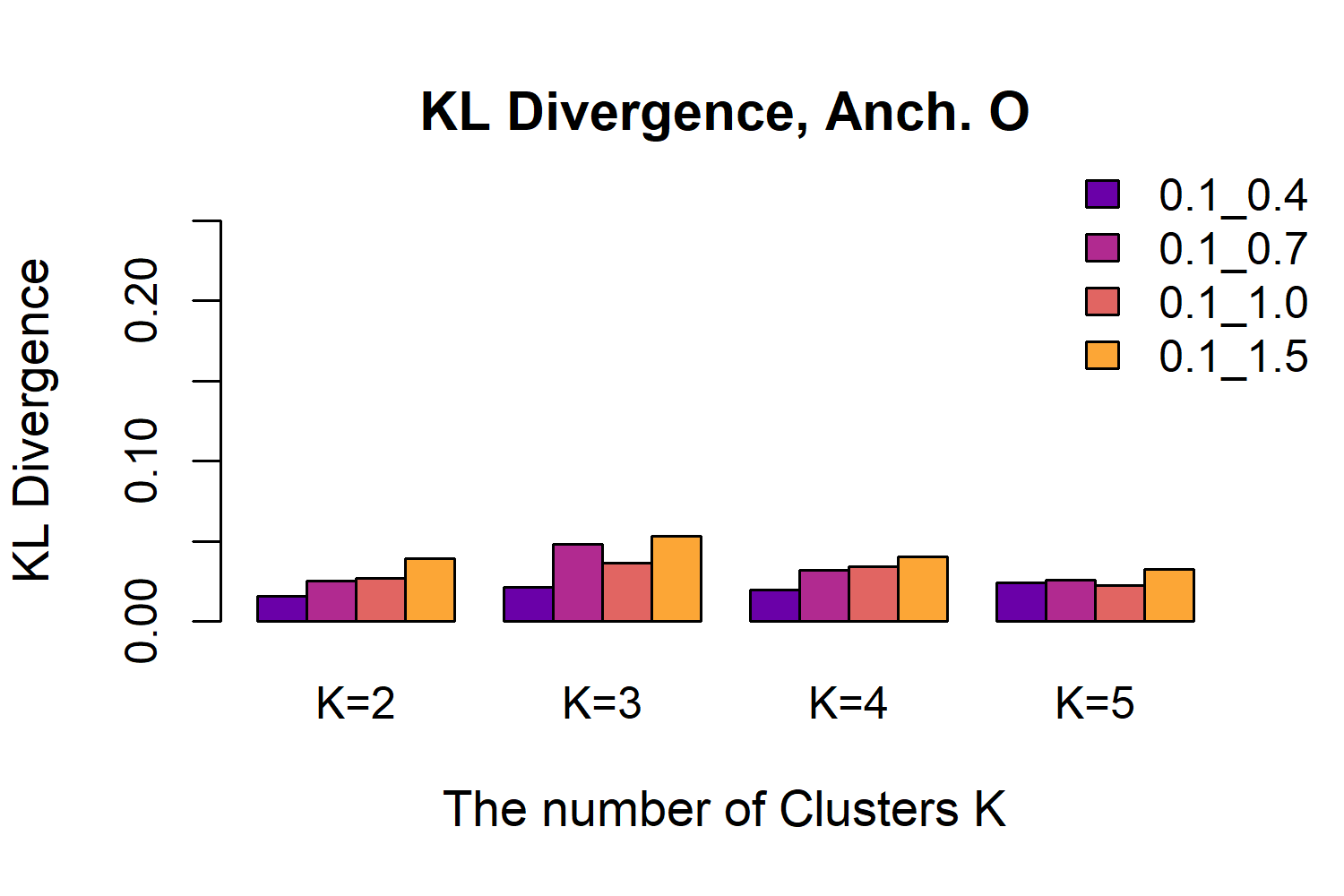}
    \end{minipage}
    \caption{ Kullback-Leibler  Divergence calculated on the $\mathbb{D}_1$ and  $\mathbb{D}_2$ for Review data. (Left) The setting used to calculate $\mathbb{D}_2$ and $\mathbb{D}_2$ corresponds to those used in testing $  H_{0} (\mathcal{G}_{\rho_1},\{ \mathcal{O}, \mathcal{G}_{\rho_2}\})$. (Right) The setting used to calculate $\mathbb{D}_2$ and $\mathbb{D}_2$ corresponds to those used in testing $H_{0} (\mathcal{O}, \{\mathcal{G}_{\rho_1}, \mathcal{G}_{\rho_2}\})$. }
    \label{fig:kl_anch_combined}
\end{figure*}


\begin{table}
    \centering
\caption{The p-values for testing the difference between $\mathcal{O}$ and $\mathcal{G}_{\rho_2}$.  For our proposed method, $\mathcal{O}$ is used as the anchor dataset. ``Anch.'' denotes the result from our proposed method with the anchored dataset represented in parenthesis. }
    \label{tab:hypo6}
    \belowrulesep = 0pt
    \aboverulesep = 0pt
    \small 
    \begin{tabular}{p{1.1cm} p{0.85cm} p{1.3cm} p{1.3cm} p{1.3cm} p{1.3cm} p{1.3cm} p{1.3cm} p{1.3cm} p{1.3cm}}
        \toprule
        \multirow{2}{*}{$\rho$ pair} & \multirow{2}{*}{Dataset} & \multicolumn{8}{c}{Testing Method} \\  
        \cmidrule(lr){3-10}
        & & \multicolumn{4}{c}{Anch. ($\mathcal{G}$)} & \multirow{2}{*}{Hotelling} & \multirow{2}{*}{Nploc} & \multirow{2}{*}{Energytest} & \multirow{2}{*}{Balltest} \\            
        & & \multicolumn{1}{c}{$K=2$} & \multicolumn{1}{c}{$K=3$} & \multicolumn{1}{c}{$K=4$} & \multicolumn{1}{c}{$K=5$} & & & & \\ 
        \cmidrule(lr){1-2} \cmidrule(lr){3-6} \cmidrule(lr){7-10} 
        $(0.1, 0.4)$ 
        & Review & 0.140 & 0.288 & 0.502 & 0.398 & $0^{*}$ & $< 1e-3^{*}$ & $0.005^{*}$ & $0.005^{*}$ \\
        & CNN    & $< 1e-3^{*}$ & 0.264 & 0.403 & 0.520 & $0.005^{*}$ & $< 1e-3^{*}$ & $0.005^{*}$ & $0.005^{*}$ \\
        & Quora  & $< 1e-3^{*}$ & $< 1e-3^{*}$ & $< 1e-3^{*}$ & $< 1e-3^{*}$ & $< 1e-3^{*}$ & $< 1e-3^{*}$ & $0.005^{*}$ & $0.005^{*}$ \\
        & SQuAD2 & 0.114 & 0.758 & $< 1e-3^{*}$ & $0.004^{*}$ & $< 1e-3^{*}$ & $< 1e-3^{*}$ & $0.005^{*}$ & $0.005^{*}$ \\
        \cmidrule(lr){1-2} \cmidrule(lr){3-6} \cmidrule(lr){7-10}
        $(0.1, 0.7)$ 
        & Review & 0.277 & $< 1e-3^{*}$ & $< 1e-3^{*}$ & $< 1e-3^{*}$ & $< 1e-3^{*}$ & $< 1e-3^{*}$ & $0.005^{*}$ & $0.005^{*}$ \\
        & CNN    & $0.005^{*}$ & 0.931 & 0.064 & $< 1e-3^{*}$ & $< 1e-3^{*}$ & $< 1e-3^{*}$ & $0.005^{*}$ & $0.005^{*}$ \\
        & Quora  & $0.020^{*}$ & $< 1e-3^{*}$ & $< 1e-3^{*}$ & $< 1e-3^{*}$ & $< 1e-3^{*}$ & $< 1e-3^{*}$ & $0.005^{*}$ & $0.005^{*}$ \\
        & SQuAD2 & $0.001^{*}$ & $0.024^{*}$ & $< 1e-3^{*}$ & $< 1e-3^{*}$ & $< 1e-3^{*}$ & $< 1e-3^{*}$ & $0.005^{*}$ & $0.005^{*}$ \\
        \cmidrule(lr){1-2} \cmidrule(lr){3-6} \cmidrule(lr){7-10}
        $(0.1, 1.0)$ 
        & Review & 0.440 & $0.002^{*}$ & $0.002^{*}$ & $< 1e-3^{*}$ & $< 1e-3^{*}$ & $< 1e-3^{*}$ & $0.005^{*}$ & $0.005^{*}$ \\
        & CNN    & $0.005^{*}$ & 0.094 & $0.031^{*}$ & $< 1e-3^{*}$ & $0^{*}$ & $< 1e-3^{*}$ & $0.005^{*}$ & $0.005^{*}$ \\
        & Quora  & 0.078 & $< 1e-3^{*}$ & 0.273 & $< 1e-3^{*}$ & $0^{*}$ & $< 1e-3^{*}$ & $0.005^{*}$ & $0.005^{*}$ \\
        & SQuAD2 & 0.263 & 0.134 & $< 1e-3^{*}$ & $0.026^{*}$ & $< 1e-3^{*}$ & $< 1e-3^{*}$ & $0.005^{*}$ & $0.005^{*}$ \\
        \cmidrule(lr){1-2} \cmidrule(lr){3-6} \cmidrule(lr){7-10}
        $(0.1, 1.5)$ 
        & Review & $< 1e-3^{*}$ & $< 1e-3^{*}$ & 0.298 & $< 1e-3^{*}$ & $0^{*}$ & $< 1e-3^{*}$ & $0.005^{*}$ & $0.005^{*}$ \\
        & CNN    & 0.625 & 0.419 & $0.004^{*}$ & $< 1e-3^{*}$ & $< 1e-3^{*}$ & $< 1e-3^{*}$ & $0.005^{*}$ & $0.005^{*}$ \\
        & Quora  & $< 1e-3^{*}$ & $< 1e-3^{*}$ & $< 1e-3^{*}$ & 0.949 & $0^{*}$ & $< 1e-3^{*}$ & $0.005^{*}$ & $0.005^{*}$ \\
        & SQuAD2 & $< 1e-3^{*}$ & 0.854 & 0.057 & $0.009^{*}$ & $< 1e-3^{*}$ & $< 1e-3^{*}$ & $0.005^{*}$ & $0.005^{*}$ \\
        \cmidrule(lr){1-2} \cmidrule(lr){3-6} \cmidrule(lr){7-10}
    \bottomrule
    \end{tabular}
\end{table}


\begin{figure*}[t]
    \centering
    \begin{minipage}[t]{0.48\textwidth}
        \centering
        \includegraphics[width=\textwidth]{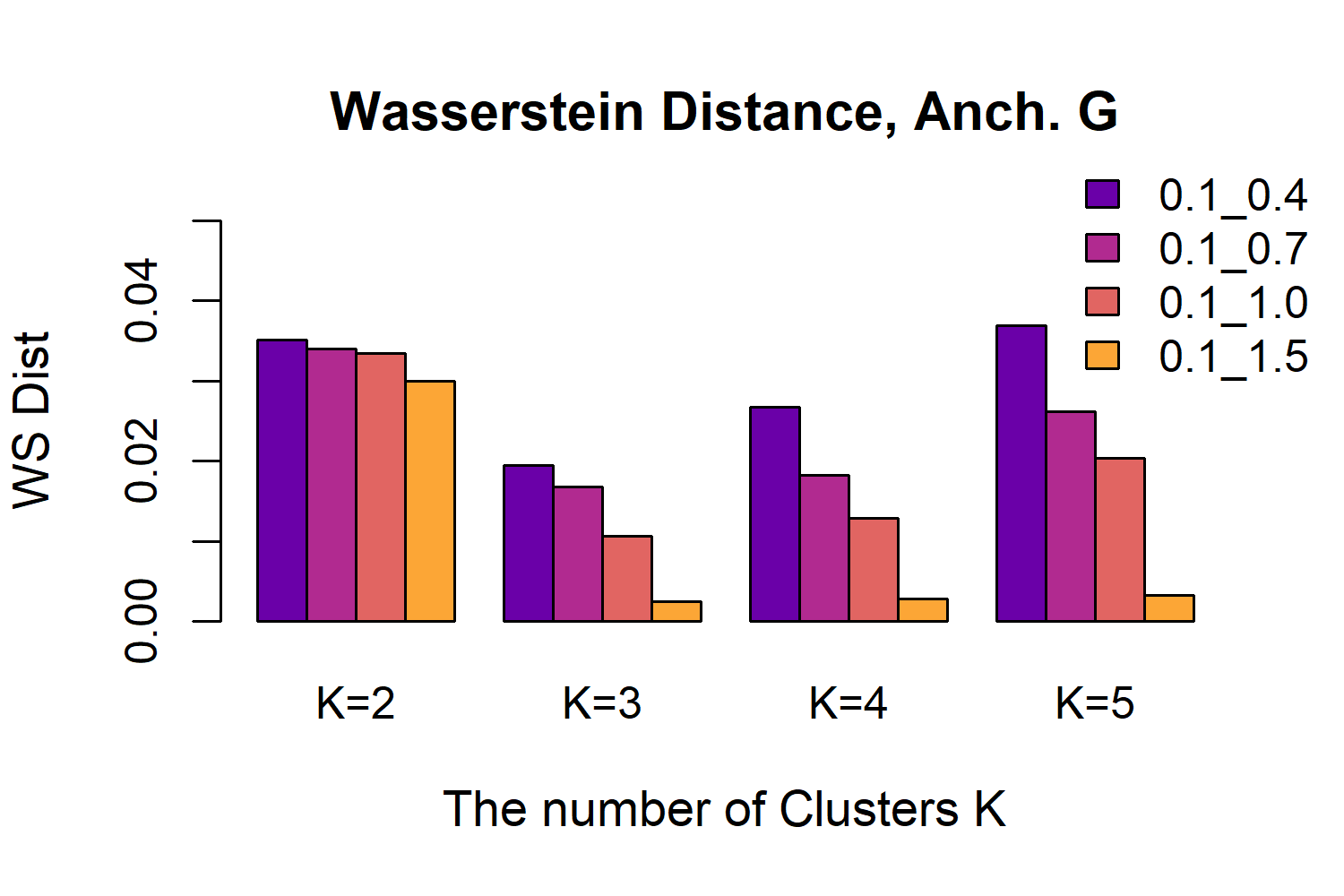}
    \end{minipage}%
    \hfill
    \begin{minipage}[t]{0.48\textwidth}
        \centering
        \includegraphics[width=\textwidth]{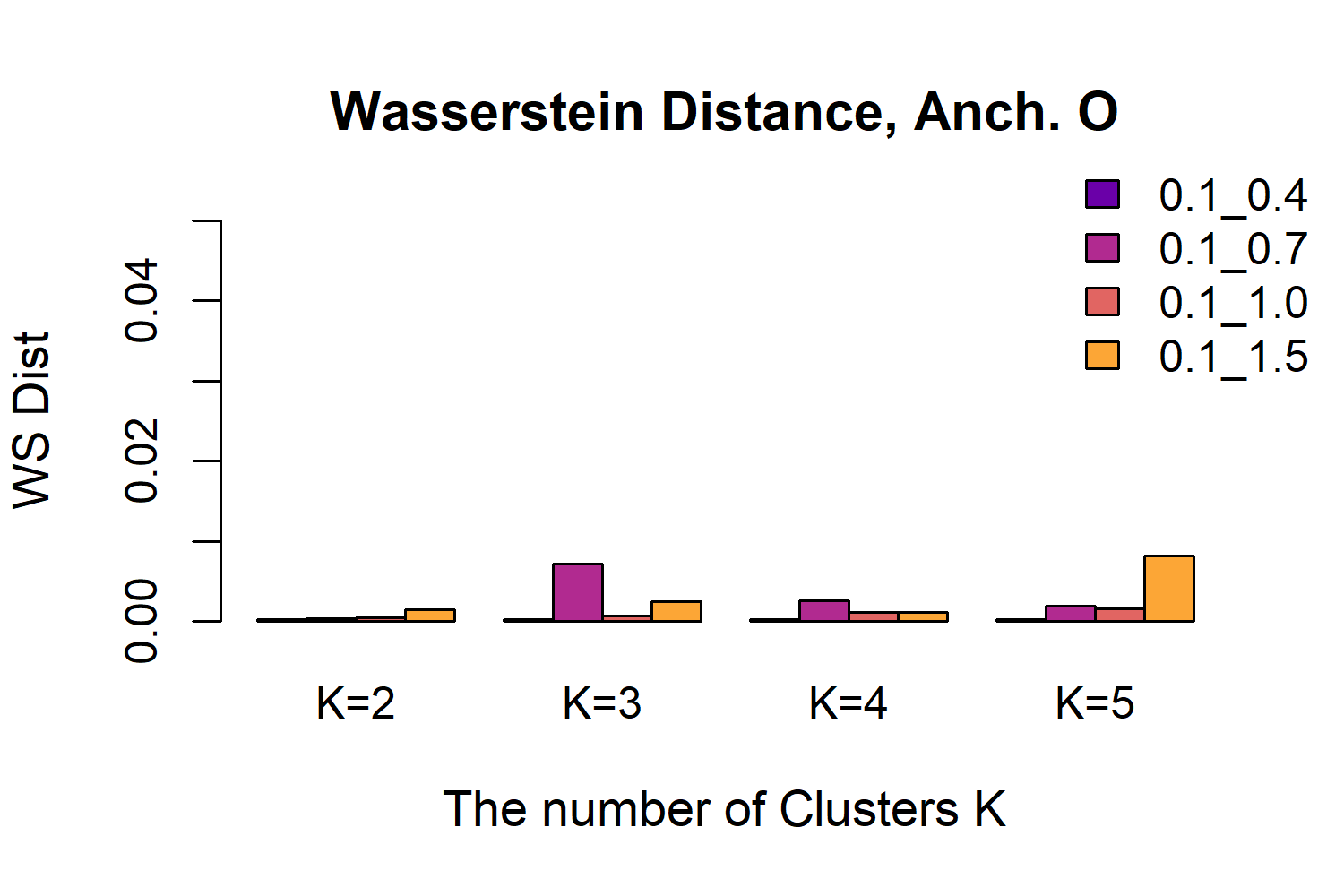}
    \end{minipage}
    \caption{ Wasserstein Distance calculated on the $\mathbb{D}_1$ and  $\mathbb{D}_2$ for Review data. (Left) The setting used to calculate $\mathbb{D}_2$ and $\mathbb{D}_2$ corresponds to those used in testing $  H_{0} (\mathcal{G}_{\rho_1},\{ \mathcal{O}, \mathcal{G}_{\rho_2}\})$. (Right) The setting used to calculate $\mathbb{D}_2$ and $\mathbb{D}_2$ corresponds to those used in testing $H_{0} (\mathcal{O}, \{\mathcal{G}_{\rho_1}, \mathcal{G}_{\rho_2}\})$. }
    \label{fig:ws_anch_combined}
\end{figure*}


\subsection{Analysis on Question 2 \label{subsec:anal_q2}}


In this section, we perform analysis on testing the hypotheses $  H_{0} (\mathcal{G}_{\rho_1},\{ \mathcal{O}, \mathcal{G}_{\rho_2}\})$ and  $H_{0} (\mathcal{O}, \{\mathcal{G}_{\rho_1}, \mathcal{G}_{\rho_2}\})$ for varying temperature parameter $\rho$, discussed in Section \ref{subsec:hyptwo}. Specifically, we set the temperature parameter \(\rho_1\) to 0.1, while \(\rho_2\) takes the values 0.4, 0.7, 1.0, and 1.5.

Table~\ref{tab:hypo5} shows the test results for the null hypothesis \(H_{0} (\mathcal{G}_{\rho_1}, \{ \mathcal{O}, \mathcal{G}_{\rho_2}\})\). The null hypothesis is rejected in most cases, indicating that the degree of variability between the pairs \(( \mathcal{G}_{\rho_1}, \mathcal{O})\) and \(( \mathcal{G}_{\rho_1}, \mathcal{G}_{\rho_2})\) is significant, with the exception of the SQuAD2 dataset for our proposed testing method when \(\rho_2 = 1.5\). Notably, the null hypothesis tends to be accepted for the SQuAD2 dataset as \(\left|\rho_1 - \rho_2\right|\) increases. This may suggest that the variability in human-authored texts and LLM-generated texts becomes relatively closer as larger temperature parameter introduces more variability in the LLM-generated texts.
On the other hand, 
Table~\ref{tab:hypo6} exhibits that  the null  $H_{0} (\mathcal{O}, \{\mathcal{G}_{\rho_1}, \mathcal{G}_{\rho_2}\})$  is accepted relatively often in our proposed testing, which may imply that the difference between two pairs $(  \mathcal{O}, \ \mathcal{G}_{\rho_1})$ and $(\mathcal{G}_{\rho_1}, \ \mathcal{G}_{\rho_2})$ are often insignificant. 
This may be due to the fact that the gap between human-authored texts and LLM-generated texts with \(\rho_1 = 0.1\) is already noticeable. As a result, the relative difference between the two pairs \((\mathcal{O}, \ \mathcal{G}_{\rho_1})\) and \((\mathcal{O}, \ \mathcal{G}_{\rho_2})\), anchored at \(\mathcal{O}\), does not show a significant difference, even though \(\rho_2\) increases.

Additionally, two statistical distances of Kullback-Leibler divergence and Wasserstein distance, between \(\mathbb{D}_1\) and \(\mathbb{D}_2\) which are used to test $  H_{0} (\mathcal{G}_{\rho_1},\{ \mathcal{O}, \mathcal{G}_{\rho_2}\})$ and  $H_{0} (\mathcal{O}, \{\mathcal{G}_{\rho_1}, \mathcal{G}_{\rho_2}\})$
are presented in Figures \ref{fig:kl_anch_combined} and \ref{fig:ws_anch_combined}, respectively.
The left panels in both figures show that the distances tend to decrease as $\rho_2$, the temperature parameter of the non-anchor dataset \(\mathcal{G}_{\rho_2}\), increases. This finding indicates that as the LLM-generated text gains more variability, the degree of discrepancy between \((\mathcal{G}_{\rho_1}, \mathcal{G}_{\rho_2})\) and that between \((\mathcal{G}_{\rho_1}, \mathcal{O})\) become closer with fixed \(\rho_1 = 0.1\), which aligns with our findings in hypothesis testing results in Table \ref{tab:hypo5}.
The right panels in both figures show smaller distance values compared to the left panels, which is consistent with our hypothesis testing results in Table \ref{tab:hypo6}. Specifically, increasing the gap between \(\rho_1\) and \(\rho_2\) does not substantially affect the relative difference between human-authored and LLM-generated texts. These two observations suggest that human-authored texts are notably distinct from LLM-generated texts.

\section{Discussion}

In this study, we explored the differences in community structures between human-authored text and LLM-generated texts. Our investigation was based on text datasets that consist of LLM paraphrase results when human-authored text is inputted, thus creating a paired structure with the original inputs. 
By leveraging the paired structure of the datasets, we proposed a hypothesis testing procedure that addresses the challenges of directly measuring distributional differences. This is achieved by establishing an anchor set that reflects the distributional differences of the other sets we wish to compare.

Our proposed method demonstrated that the original human text input and its LLM-generated paraphrase exhibit difference in community structures, while LLM-generated texts tend to be relatively similar to each other. The observed gap between human-authored text and LLM-generated text suggests that future advancements in sophisticated language processing and contextual understanding may be necessary.

Additionally, there are limited methods and metrics available for quantitatively assessing the performance of LLMs.
Our proposed testing method contributes to this area, though it has limitations, such as capturing only indirect evidence of disparity, which may reduce its detection power and applicability in settings like paired data. Developing such methods opens up an intriguing research area.

\FloatBarrier

\printbibliography

\end{document}